\documentclass[final]{cvpr}

\usepackage{times}
\usepackage{epsfig}
\usepackage{graphicx}
\usepackage{amsmath}
\usepackage{amssymb}

\usepackage{caption}
\usepackage{subcaption}
\usepackage{algorithm}
\usepackage{algorithmic}
\usepackage{multicol}
\usepackage{booktabs}
\usepackage{soul}
\usepackage[section]{placeins}
\usepackage{balance}

\newcommand\blfootnote[1]{%
  \begingroup
  \renewcommand\thefootnote{}\footnote{#1}%
  \addtocounter{footnote}{-1}%
  \endgroup
}

\usepackage[pagebackref=true,breaklinks=true,colorlinks,bookmarks=false]{hyperref}

\newtheorem{proposition}{Proposition}[section]

\def\cvprPaperID{3798} 
\def\confYear{CVPR 2021}

\begin{document}

\title{Deep Stable Learning for Out-Of-Distribution Generalization}

\author{Xingxuan Zhang, Peng Cui*, Renzhe Xu, Linjun Zhou, Yue He, Zheyan Shen\\
Department of Computer Science, Tsinghua University, Beijing, China\\
{\tt\small {xingxuanzhang@hotmail.com, cuip@tsinghua.edu.cn, xrz199721@gmail.com,}} \\
{\tt\small {\{zhoulj16, heyue18, shenzy17\}@mails.tsinghua.edu.cn}}}

\maketitle

\pagestyle{empty}  
\thispagestyle{empty} 

\begin{abstract}
  Approaches based on deep neural networks have achieved striking performance when testing data and training data share similar distribution, but can significantly fail otherwise.
  Therefore, eliminating the impact of distribution shifts between training and testing data is crucial for building performance-promising deep models. 
Conventional methods assume either the known heterogeneity of training data (e.g. domain labels) or the approximately equal capacities of different domains. 
In this paper, we consider a more challenging case where neither of the above assumptions holds. 
We propose to address this problem by removing the dependencies between features via learning weights for training samples, which helps deep models get rid of spurious correlations and, in turn, concentrate more on the true connection between discriminative features and labels. 
Extensive experiments clearly demonstrate the effectiveness of our method on multiple distribution generalization benchmarks compared with state-of-the-art counterparts. 
Through extensive experiments on distribution generalization benchmarks including PACS, VLCS, MNIST-M, and NICO, we show the effectiveness of our method compared with state-of-the-art counterparts. 

\end{abstract}

\blfootnote{*Corresponing author, also with Beijing Key Lab of Networked Multimedia}

\section{Introduction} \label{para:intro}

\begin{figure}[th]
    \centering
    \includegraphics[width=\linewidth]{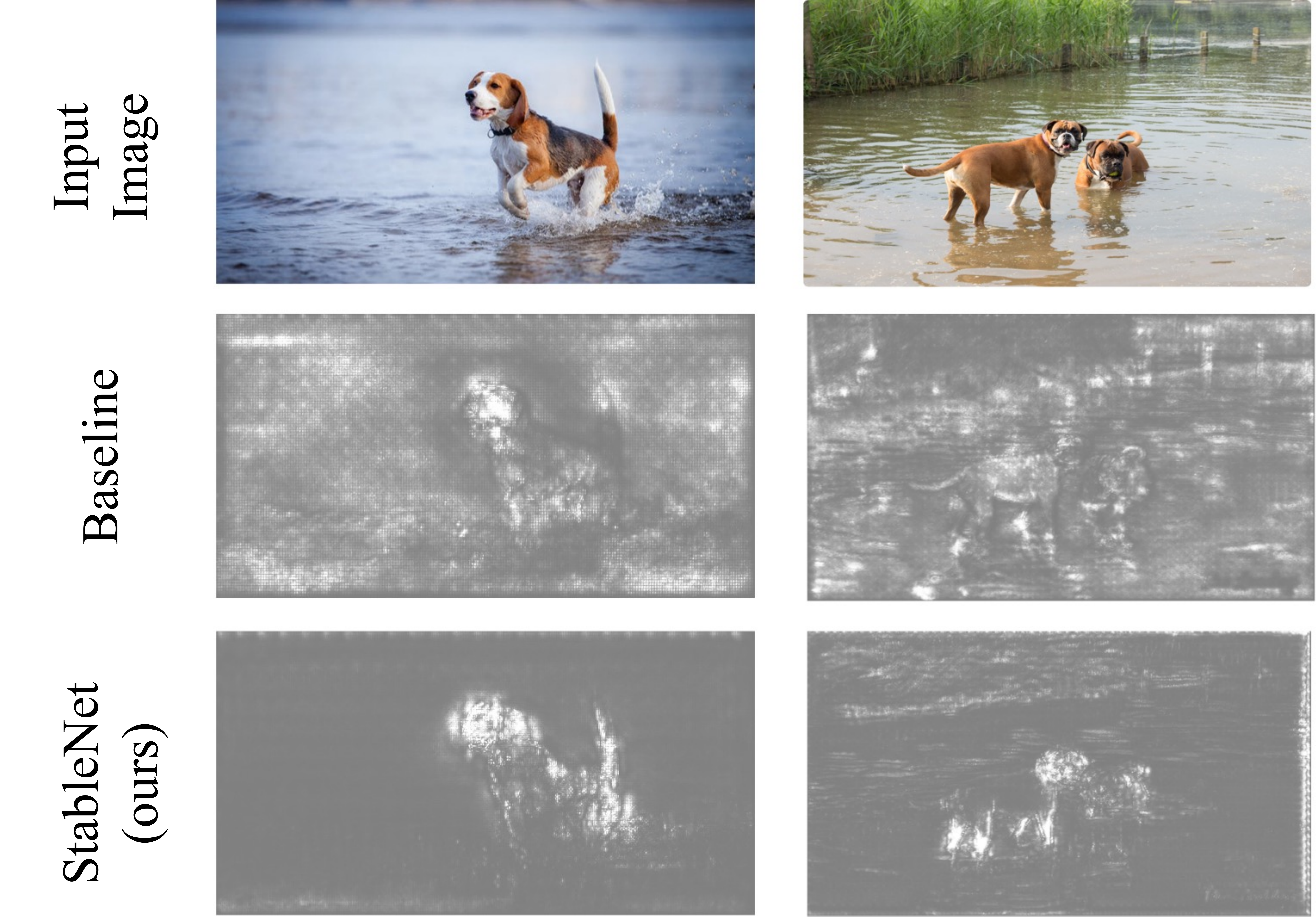}
    \caption{Visualization of saliency maps produced by the vanilla ResNet-18 model and StableNet when most of the training images containing dogs in the water.
    The lightness of the saliency map indicates how much attention that the models pay on particular area of the input image (i.e. lighter area plays a more crucial role for the prediction than the darker area).
    Due to the spurious correlation, the ResNet-18 model tends to focus on both dogs and the water while our model focuses mostly on dogs.}
    \label{fig:intro}
\end{figure}

Many machine learning approaches tend to exploit subtle statistical correlations existing in the training distribution for predictions which have been shown to be effective under the I.I.D. hypothesis, i.e., testing and training data is independently sampled from the identical distribution.
In real cases, however, such a hypothesis can hardly be satisfied due to the complex generation mechanism of real data such as data selection biases, confounding factors, or other peculiarities \cite{2019A, 2017One, 2017Exploring, 2019Do, 2019Benchmarking}.
The testing distribution may incur uncontrolled and unknown shifts from the training distribution, which makes most machine learning models fail to make trustworthy predictions \cite{arjovsky2019invariant, shen2020stable}. 
To address this issue, out-of-distribution (OOD) generalization is proposed for improving the generalization ability of models under distribution shifts \cite{2019Test, 2020Out}.

Essentially, when there incurs a distribution shift, the accuracy drop of current models is mainly caused by the spurious correlation between the irrelevant features (i.e. the features that are irrelevant to a given category, such as features of context, figure style, etc.) and category labels, and this kind of spurious correlations are intrinsically caused by the subtle correlations between irrelevant features and relevant features (i.e. the features that are relevant to a given category) \cite{lake2017building, marcus2018deep, lopez2017discovering, arjovsky2019invariant}. 
Taking the recognition task of `dog' category as an example, as depicted in Figure \ref{fig:intro}, if dogs are in the water in most training images, the visual features of dogs and water would be strongly correlated, thus leading to the spurious correlation between visual features of water with the label `dog'. 
As a result, when encountering images of dogs without water, or other objects (such as cats) with water, the model is prone to produce false predictions.

Recently, such distribution (domain) shift problems have been intensively studied in the \textsl{domain generalization (DG)} literature \cite{muandet2013domain,  ghifary2015domain,khosla2012undoing, wang2020learning,li2017deeper,li2019episodic}. 
The basic idea of DG is to divide a category into multiple domains so that irrelevant features vary across different domains while relevant features remain invariant \cite{khosla2012undoing, li2018domain, motiian2017unified}. 
Such training data makes it possible for a well-designed model to learn the invariant representations across domains and inhibit the negative effect from irrelevant features, leading to better generalization ability under distribution shifts. 
Some pioneering methods require clear and significant heterogeneity, namely that the domains are manually divided and labeled \cite{wang2017select, ganin2016domain, ratner2017learning, ding2017deep, niu2015multi}, which cannot be always satisfied in real applications. 
More recently, some methods are proposed to implicitly learn latent domains from data \cite{qiao2020learning, matsuura2020domain, wang2019learning}, but they implicitly assume that the latent domains are balanced, meaning that the training data is formed by balanced sampling from latent domains. 
In real cases, however, the assumption of domain balance can be easily violated, leading to the degeneration of these methods. 
This is also empirically validated in our experiments as shown in Section \ref{gen_inst}. 

Here we consider a more realistic and challenging setting where the domains of training data are unknown and we do not implicitly assume that the latent domains are balanced. With this goal, a strand of research on stable learning are proposed \cite{2017On,2018Stable}. 
Given that the statistical dependence between relevant and irrelevant features is a major cause of model crash under distribution shifts, they propose to realize out-of-distribution generalization by decorrelating the relevant and irrelevant features. 
Since there is no extra supervision for separating relevant features from irrelevant features, a conservative solution is to decorrelate all features. 
Recently, this notion has been demonstrated to be effective in improving the generalization ability of linear models. 
\cite{kuang2020stable} proposes a sample weighting approach with the goal of decorrelating input variables, and \cite{shen2020stable} theoretically proves why such sample weighting can make a linear model produce stable predictions under distribution shifts.
But they are all developed under the constraints of linear frameworks.
When extending these ideas into deep models to tackle more complicated data types like images, we confront two main challenges. First, the complex non-linear dependencies among features are much more difficult to be measured and eliminated than the linear ones.
Second, the global sample weighting strategy in these methods requires excessive storage and computational cost in deep models, which is infeasible in practice.

To address these two challenges, we propose a method called \textbf{StableNet}.
In terms of the first challenge, we propose a novel nonlinear feature decorrelation approach based on Random Fourier Features \cite{rahimi2008random} with linear computational complexity. 
As for the second challenge, we propose an efficient optimization mechanism to perceive and remove correlations globally by iteratively saving and reloading features and weights of the model. 
These two modules are jointly optimized in our method. 
Moreover, as shown in Figure \ref{fig:intro}, StableNet can effectively partial out the irrelevant features (i.e. water) and leverage truly relevant features for prediction, leading to more stable performances in the wild non-stationary environments.

\section{Related Works} \label{para:related_work}


\noindent \textbf{Domain Generalization}. Domain generalization (DG) considers the generalization capacities to unseen domains of deep models trained with multiple source domains. 
A common approach is to extract domain-invariant features over multiple source 
domains \cite{ghifary2015domain,khosla2012undoing,li2017learning,li2018domain,motiian2017unified, dou2019domain, hu2020domain, piratla2020efficient, seo2019learning, motiian2017unified} 
or to aggregate domain-specific modules \cite{mancini2018best, mancini2018robust}. 
Several works propose to enlarge the available data space with augmentation of source domains \cite{carlucci2019domain, shankar2018generalizing, volpi2018generalizing, qiao2020learning,zhou2020learning, zhou2020deep}. 
There are several approaches that exploit regularization with meta-learning \cite{li2019episodic, dou2019domain} and Invariant Risk Minimization (IRM) framework \cite{arjovsky2019invariant} for DG.
Despite the promising results of DG methods in the well-designed experimental settings, some strong assumptions such as the manually divided and labeled domains and the balanced sampling process from each domain actually hinder the DG methods from real applications.    

\noindent \textbf{Feature Decorrelation}. As the correlations between features affect or even impair the model prediction, several works have focused on remove such correlation in the training process.
Some pioneering works based on Lasso framework \cite{takada2018IILasso, chen2013uncorrelated} propose to decorrelate features by adding a regularizer that imposes the highly correlated features not to be selected simultaneously.
Recently, several works theoretically bridge the connections between correlation and model stability under misspecification \cite{shen2020stable, kuang2020stable}, and propose to address such a problem via a sample reweighting scheme.
However, the above methods are all developed under linear frameworks which can not handle complex data types such as images and videos in computer vision applications. More related works and discussions are in Appendix A.
\section{Sample Weighting for Distribution Generalization}
\label{headings}

\begin{figure*}[ht]
    \centering
    \includegraphics[width=0.75\textwidth]{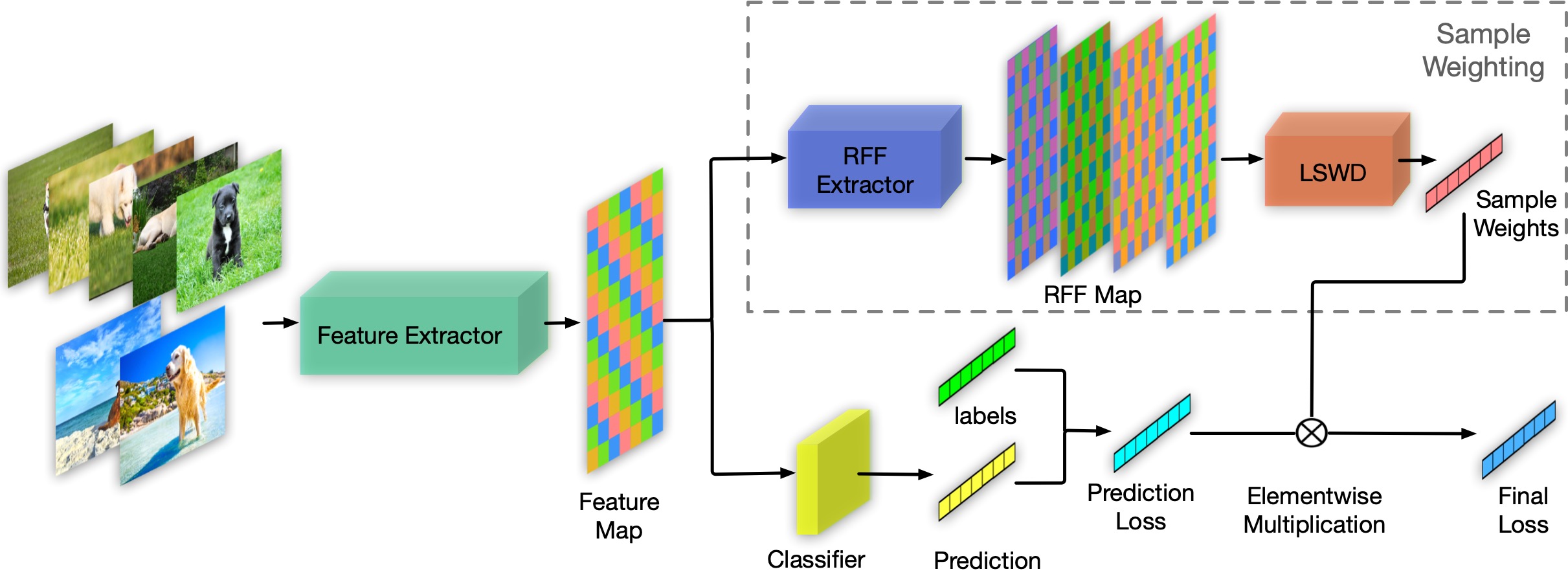}
    \vspace{-8pt}
    \caption{The overall architecture of the proposed StableNet. LSWD refers to \textit{learning sample weighting for decorrelation} as described in Section \ref{method_1}. \textit{Final loss} is used to optimized the classification network. Detailed learning procedure of StableNet is in Section \ref{method_1} and Appendix B.1.}
    \label{fig:model}
    \vspace{-8pt}
\end{figure*}

We address the distribution shifts problem by weighting samples globally to directly decorrelate all the features for every input sample, thus statistical correlations between relevant and irrelevant features are eliminated.
 Concretely, StableNet gets rid of both linear and non-linear dependencies between features by utilizing the characteristics of Random Fourier Features (RFF) and sample weighting. 
To adapt the global decorrelation method to modern deep models, we further propose the saving and reloading global correlation mechanism, to decrease the usage of storage and computational cost when the training data are of a large scale. 
The formulations and theoretical explanations are shown in Section \ref{method_1}. In Section \ref{method_2}, we introduce the saving and reloading global correlation method, which makes calculating correlation globally possible with deep models.   
\textbf{Notations} $\mathcal{X} \subset \mathbb{R}^{m_X}$ denotes the space of raw pixels, $\mathcal{Y} \subset \mathbb{R}^{m_Y}$ denotes the outcome space and $\mathcal{Z} \subset \mathbb{R}^{m_Z}$ denotes the representation space. $m_X$, $m_Y$, $m_Z$ are the dimensions of space $\mathcal{X}$, $\mathcal{Y}$, $\mathcal{Z}$, respectively. $f: \mathcal{X} \rightarrow \mathcal{Z}$ denotes the representation function and $g: \mathcal{Z} \rightarrow \mathcal{Y}$ denotes the prediction function. We have $n$ samples $\mathbf{X} \subset \mathbb{R}^{n \times m_X}$ with labels $\mathbf{Y} \subset \mathbb{R}^{n \times m_Y}$ and we use $\mathbf{X}_i$ and $y_i$ to denote the $i$-th sample. The representations learned by neural networks are donated as $\mathbf{Z} \subset \mathbb{R}^{n \times m_Z}$ and the $i$-th variable in the representation space is donated as $\mathbf{Z}_{:,i}$. We use $\mathbf{w} \in \mathbb{R}^n$ to denote sample weights. $\mathbf{u}$ and $\mathbf{v}$ are Random Fourier Features mapping functions. 

\subsection{Sample weighting with RFF}
\label{method_1}


\paragraph{Independence testing statistics} To eliminate the dependence between any pair of features $\mathbf{Z}_{:, i}$ and $\mathbf{Z}_{:, j}$ in the representation space, we introduce hypothesis testing statistics that measures the independence between random variables. Suppose there are two one-dimensional random variables $A, B$ (Here we use $A$ and $B$ to represent random variables instead of $\mathbf{Z}_{:, i}$ and $\mathbf{Z}_{:, j}$ for simplicity of notation.) and we sample $(A_1, A_2, \dots A_n)$ and $(B_1, B_2, \dots B_n)$ from the distribution of $A$ and $B$, respectively. 
The main problem is how relevant these two variables are based on the samples.




Consider a measurable, positive definite kernel $k_A$ on the domain of random variable $A$ and the corresponding RKHS is denoted by $\mathcal{H}_A$. If $k_B$ and $\mathcal{H}_B$ are similarly defined, the cross-covariance operator $\Sigma_{AB}$ \cite{fukumizu2004dimensionality} from $\mathcal{H}_B$ to $\mathcal{H}_A$ is as follows: 
\begin{equation}
    \begin{aligned}
        & \langle h_A,\Sigma_{AB}h_B\rangle \\
        = & \mathbb{E}_{AB}\left[h_A(A)h_B(B)\right] - \mathbb{E}_A[h_A(A)]\mathbb{E}_B[h_B(B)]
    \end{aligned}
\end{equation}
for all $h_A \in \mathcal{H}_A$ and $h_B \in \mathcal{H}_B$. Then, the independence can be determined by the following proposition \cite{fukumizu2008kernel}.

\begin{proposition}
    If the product $k_A k_B$ is characteristic, $\mathbb{E}[k_{A}(A,A)] < \infty$ and $\mathbb{E}[k_{B}(B,B)] < \infty$, we have
    \begin{equation}
        \Sigma_{AB} = 0 \Longleftrightarrow A \perp B
    \end{equation}
\end{proposition}

Hilbert-Schmidt Independence Criterion (HSIC) \cite{gretton2008kernel}, which requires that the squared Hilbert-Schmidt norm of $\Sigma_{AB}$ should be zero, can be applied as a criterion to supervise feature decorrelation \cite{bahng2019learning}.
However, the calculation of HSIC requires noticeable computational cost which grows as the batch size of training data increases, so it is inapplicable to training deep models on large datasets. More approaches of independence test are discussed in Appendix B.2.
Actually, Frobenius norm corresponds to the Hilbert-Schmidt norm in Euclidean space \cite{strobl2019approximate}, so that the independent testing statistic can be based on Frobenius norm.

Let the partial cross-covariance matrix be:

\begin{small}
\begin{equation}\label{eq:cross-covariance}
    \begin{aligned}
        \hat{\Sigma}_{AB} = \frac{1}{n - 1} \sum_{i=1}^n \bigg[ & \bigg(\mathbf{u}(A_i) - \frac{1}{n}\sum_{j=1}^n \mathbf{u}(A_j)\bigg)^T \cdot \\
        & \bigg(\mathbf{v}(B_i) - \frac{1}{n}\sum_{j=1}^n \mathbf{v}(B_j)\bigg)\bigg],
    \end{aligned}
\end{equation}
\end{small}
where
\begin{equation}\label{eq:rff_transform}
    \begin{aligned}
        \mathbf{u}(A) & = \left(u_1(A), u_2(A), \dots u_{n_A}(A)\right), u_j(A) \in \mathcal{H}_\text{RFF}, \forall j,\\
        \mathbf{v}(B) & = \left(v_1(B), v_2(B), \dots v_{n_B}(B)\right), v_j(B) \in \mathcal{H}_\text{RFF}, \forall j.
    \end{aligned}
\end{equation}
Here we sample $n_A$ and $n_B$ functions from $\mathcal{H}_\text{RFF}$ respectively and $\mathcal{H}_\text{RFF}$ denotes the function space of Random Fourier Features with the following form

\begin{equation}
    \begin{aligned}
        \mathcal{H}_\text{RFF} = \big\{ & h: x \rightarrow \sqrt{2}\cos(\omega x + \phi) \mid \\
        & \omega \sim N(0, 1), \phi \sim \text{Uniform}(0, 2 \pi)\big\},
    \end{aligned}
\end{equation}
\textit{i.e.} $\omega$ is sampled from the standard Normal distribution and $\phi$ is sampled from the Uniform distribution. Then, the independence testing statistic $I_{AB}$ is defined as the Frobenius norm of the partial cross-covariance matrix, \textit{i.e.}, $I_{AB} =  \left\Vert\hat{\Sigma}_{AB}\right\Vert_F^2$.

Notice that $I_{AB}$ is always non-negative. As $I_{AB}$ decreases to zero, the two variables $A$ and $B$ tends to be independent. Thus $I_{AB}$ can effectively measure the independence between random variables.
The accuracy of independence test grows as $n_A$ and $n_B$ increase. Empirically, setting both $n_A$ and $n_B$ to $5$ is solid enough to judge the independence of random variables \cite{strobl2019approximate}.

\paragraph{Learning sample weights for decorrelation} Inspired by \cite{kuang2020stable}, we propose to eliminate the dependence between features in the representation space via sample weighting and measure general independence via RFF.

We use $\mathbf{w} \in \mathbb{R}^n_+$ to denote the sample weights and $\sum_{i=1}^n w_i = n$. After weighting, the partial cross-covariance matrix for random variables $A$ and $B$ in Equation \ref{eq:cross-covariance} can be calculated as follows:

\begin{small}
\begin{equation}
\label{eqn:sigmaabw}
    \begin{aligned}
        \hat{\Sigma}_{AB;\mathbf{w}} = \frac{1}{n - 1} \sum_{i=1}^n \bigg[ & \bigg(w_i\mathbf{u}(A_i) - \frac{1}{n}\sum_{j=1}^n w_j\mathbf{u}(A_j)\bigg)^T \cdot \\
        & \bigg(w_i\mathbf{v}(B_i) - \frac{1}{n}\sum_{j=1}^n w_j\mathbf{v}(B_j)\bigg)\bigg].
    \end{aligned}
\end{equation}
\end{small}
Here $\mathbf{u}$ and $\mathbf{v}$ are the RFF mapping functions explained in Equation \ref{eq:rff_transform}. StableNet targets independence between any pair of features.
 Specifically, for feature $\mathbf{Z}_{:,i}$ and $\mathbf{Z}_{:,j}$, the corresponding partial cross-covariance matrix should be $\left\Vert\hat{\Sigma}_{Z_{:,i}Z_{:, j};\mathbf{w}}\right\Vert_F^2$, shown in Equation \ref{eqn:sigmaabw}. We propose to optimize $\mathbf{w}$ by

\begin{equation}\label{eq:learning_weight}
    \mathbf{w^{*}} = \underset{\mathbf{w} \in \Delta_n}{\arg \min} \sum_{1 \le i < j \le m_Z} \left\Vert\hat{\Sigma}_{\mathbf{Z}_{:,i}\mathbf{Z}_{:, j};\mathbf{w}}\right\Vert_F^2,
\end{equation}
where $\Delta_n = \left\{\mathbf{w} \in \mathbb{R}^n_+ \mid \sum_{i=1}^n w_i = n\right\}$. Hence, weighting training samples with the optimal $\mathbf{w}^*$ can mitigate the dependence between features to the greatest extent

Generally, our algorithm iteratively optimize sample weights $\mathbf{w}$, representation function $f$, and prediction function $g$ as follows:
\begin{equation} \label{eq:overall}
    \begin{aligned}
        f^{(t+1)}, g^{(t+1)} = & \underset{f, g}{\arg \min} \sum_{i=1}^n w^{(t)}_i L(g(f(\mathbf{X}_i)), y_i), \\
        \mathbf{w}^{(t+1)} = & \underset{\mathbf{w} \in \Delta_n}{\arg \min} \sum_{1 \le i < j \le m_Z} \left\Vert\hat{\Sigma}_{\mathbf{Z}^{(t+1)}_{:,i}\mathbf{Z}^{(t+1)}_{:, j};\mathbf{w}}\right\Vert_F^2. \\
    \end{aligned}
\end{equation}
where $\mathbf{Z}^{(t+1)}=f^{(t+1)}(\mathbf{X})$, $L(\cdot, \cdot)$ represents the cross entropy loss function and $t$ represents the time stamp. Initially, $\mathbf{w}^{(0)} = (1, 1, \dots, 1)^T$.

\subsection{Learning sample weights globally}
\label{method_2}
Equation \ref{eq:overall} requires a specific weight learned for each sample. However, in practice, especially for deep learning tasks, it requires enormous storage and computational cost to learn sample weights globally.
 Moreover, with SGD for optimization, only part of the samples are observed in each batch, hence global weights for all samples cannot be learned. In this part, we propose a saving and reloading method, which merges and saves features and sample weights encountered in the training phase and reloads them as global knowledge of all the training data to optimize sample weights. 

For each batch, the features used to optimize the sample weights are generated as follows:
\begin{equation}\label{eq:concat_w}
    \begin{aligned}
        \mathbf{Z}_{O} & = \text{Concat}\left(\mathbf{Z}_{G1}, \mathbf{Z}_{G2},\cdot\cdot\cdot, \mathbf{Z}_{Gk}, \mathbf{Z}_{L} \right),\\
        \mathbf{w}_{O} & = \text{Concat}\left(\mathbf{w}_{G1}, \mathbf{w}_{G2},\cdot\cdot\cdot, \mathbf{w}_{Gk}, \mathbf{w}_{L} \right).
    \end{aligned}
\end{equation}
Here we slightly abuse the notation $\mathbf{Z}_{O}$ and $\mathbf{w}_{O}$ to mean the features and weights used to optimize the new sample weights, respectively, $\mathbf{Z}_{G1},\cdot\cdot\cdot,\mathbf{Z}_{Gk}$, $\mathbf{w}_{G1},\cdot\cdot\cdot,\mathbf{w}_{Gk}$ are global features and weights, which are updated at the end of each batch and represent global information of the whole training dataset. $\mathbf{Z}_{L}$ and $\mathbf{w}_{L}$ are features and weights in the current batch, representing the local information. The operation for merging all features in Equation \ref{eq:concat_w} is the concatenating operation along samples, \textit{i.e.} if the batch size is $B$, $\mathbf{Z}_O$ is a matrix of size $((k+1)B) \times m_Z$ and $\mathbf{w}_O$ is a $((k+1)B)$-dimensional vector. In this way, we reduce the storage and the computational cost from $O(N)$ to $O(kB)$. While training for each batch, we keep $\mathbf{w}_{Gi}$ fixed and only $\mathbf{w}_L$ is learnable under Equation \ref{eq:overall}. At the end of each iteration of training, we fuse the global information $(\mathbf{Z}_{Gi}, \mathbf{w}_{Gi})$ and the local information $(\mathbf{Z}_L, \mathbf{w}_L)$ as follows:
\begin{equation}\label{eq:save_w}
    \begin{aligned}
        \mathbf{Z}_{Gi}' & = \alpha_{i}\mathbf{Z}_{Gi} + (1 - \alpha_i)\mathbf{Z}_{L}, \\
        \quad \quad \mathbf{w}_{Gi}' & = \alpha_{i}\mathbf{w}_{Gi} + (1 - \alpha_i)\mathbf{w}_{L}.
    \end{aligned}
\end{equation}
Here for each group of global information $(\mathbf{Z}_{Gi}, \mathbf{w}_{Gi})$, we use $k$ different smoothing parameters $\alpha_i$ for considering both long-term memory ($\alpha_i$ is large) and short-term memory ($\alpha_i$ is small) in global information and $k$ indicates that the presaved features are $k$ times of that of original features. Finally, we substitute all $(\mathbf{Z}_{Gi}, \mathbf{w}_{Gi})$ with $(\mathbf{Z}_{Gi}', \mathbf{w}_{Gi}')$ for the next batch.

In the training phase, we iteratively optimize sample weights and model parameters with Equation \ref{eq:overall}. In the inference phase, the predictive model directly conduct prediction without any calculation of sample weights. The detailed procedure of our method is shown in Appendix B.1.
\section{Experiments}
\label{gen_inst}

\begin{table*}[ht]
    \centering
    \caption{Results of the \textsl{unbalanced} setting on PACS and VLCS. We reimplement the methods that require no domain labels on PACS and VLCS with ResNet18 \cite{he2016deep} which is pretrained on ImageNet \cite{deng2009imagenet} as the backbone network for all the methods. The reported results are average over three repetitions of each run. The title of each column indicates the name of the domain used as target. The best results of all methods are highlighted with the bold font and the second with underscore.}
    \scalebox{0.85}{
    \begin{tabular}{c|cccc|c|cccc|c}
        \toprule
         & \multicolumn{5}{c|}{PACS} & \multicolumn{5}{c}{VLCS} \\
         \cmidrule{2-11} 
         & Art. & Cartoon & Sketch & Photo & Avg. & Caltech & Labelme & Pascal & Sun & Avg. \\
        \midrule
    
        JiGen \cite{carlucci2019domain} & 72.76 & 69.21 & 64.90 & 91.24 & 74.53 & 85.20 & 59.73 & 62.64 & \underline{50.59} & \underline{64.54} \\
        M-ADA \cite{qiao2020learning} & 61.53 & 68.76 & 58.49 & 83.21 & 68.00 & 70.29 & 55.44 & 49.96 & 37.78 & 53.37\\
        DG-MMLD  \cite{matsuura2020domain} & 64.25 & \underline{70.31} & 64.16 & 91.64 & 72.59 & 79.76 & 57.93 & 65.25 & 44.61 & 61.89 \\
        RSC \cite{huang2020self} & \underline{75.72} & 68.50 & \underline{66.10} & \underline{93.93} & \underline{76.06} & 83.82 & 59.92 & \underline{64.49} & 49.08 & 64.33 \\
        \midrule
        ResNet-18 & 68.41 & 67.32 & 65.75 & 90.22 & 72.93 & \underline{80.02} & \underline{60.21} & 58.33 & 47.59 & 61.54\\
        StableNet (ours) & \textbf{80.16}  & \textbf{74.15} & \textbf{70.10} & \textbf{94.24} & \textbf{79.66} & \textbf{88.25}  & \textbf{62.59} & \textbf{65.77} & \textbf{55.34} & \textbf{67.99} \\
        \bottomrule
    \end{tabular}}
    \label{tab:comp_dom}

\end{table*}

        
\begin{table*}[ht]
    \centering
    \caption{Results of the \textsl{unbalanced + flexible} setting on PACS, VLCS and NICO. For details about the number of runs, meaning of column titles and fonts, see Table \ref{tab:comp_dom}.}
    \label{tab:comp_dom_nonfixed}
    \scalebox{0.85}{
    \begin{tabular}{ccccccc}
        \toprule
        &  JiGen & M-ADA & DG-MMLD & RSC & ResNet-18 & StableNet (ours)\\
        \midrule
        PACS & 40.31 &30.32 & \underline{42.65} & 39.49 &39.02 & \textbf{45.14} \\
        VLCS & 76.75 &69.58 & \underline{78.96} & 74.81 &73.77 & \textbf{79.15} \\
        \midrule
        NICO &  54.42 & 40.78 & 47.18 & \underline{57.59} & 51.71 & \textbf{59.76} \\  
        \bottomrule
    \end{tabular}}
    \label{tab:comp_dom_flexible}
    \vspace{-12pt}
\end{table*}

        

        
\begin{table*}[ht]
\caption{Results of the \textsl{unbalanced + flexible + adversarial} setting on MNIST-M. Random donates each digit is blended over a randomly chosen background. DR0.5 donates that in each class, the proportion of the dominant domain in all the training data is 50\% and other notations with `DR' are similar. 
}
\label{mnist-m}
    \centering
    \scalebox{0.85}{
    \begin{tabular}{ccccccccc}
        \toprule
        \multicolumn{1}{c}{\bf Settings}  &\multicolumn{1}{c}{Random}
        &\multicolumn{1}{c}{DR0.5} &\multicolumn{1}{c}{DR0.6} &\multicolumn{1}{c}{DR0.7} &\multicolumn{1}{c}{DR0.8} &\multicolumn{1}{c}{DR0.9} &\multicolumn{1}{c}{DR0.95}
        &\multicolumn{1}{c}{Avg.}
        \\ \midrule
       
        JiGen &97.18 &\underline{94.97} &\underline{92.99} &\underline{90.64} &78.97 &68.79 &69.34 &84.70 \\
        M-ADA &95.92 &94.45 &92.29 &88.87 &85.89 &\underline{70.32} &67.08 &84.97 \\
        DG-MMLD &96.89 &94.61 &92.59 &89.72 &\textbf{88.44} &69.13 &\underline{71.39} &\underline{86.11}\\
        RSC & \underline{96.94} & 93.43 & 89.44 & 85.78 & 81.68 & 69.15 & 65.12 &83.08\\
        \midrule
        CNNs &96.93 &93.76 &91.93 &88.13 &81.48 &68.43 &66.11  &83.82\\
        StableNet (ours) &\textbf{97.35} &\textbf{95.33} &\textbf{93.49} &\textbf{91.24} &\underline{87.04} &\textbf{75.69} &\textbf{75.46} &\textbf{87.94}\\
        \bottomrule
    \end{tabular}}
    \label{tab:mnist-m}
\end{table*}

\begin{table*}[ht]
    \centering
    \caption{Results of the \textsl{classic} setting on PACS and VLCS. All the results on PACS are obtained from the original papers of these methods. We reimplement the methods that require no domain labels on VLCS since these methods are tested with AlexNet \cite{krizhevsky2012imagenet} in original papers while we adopt ResNet18 \cite{he2016deep} as the backbone network for all the methods. The methods that require domain labels are labelled with asterisk. 
    }
    \scalebox{0.85}{
    \begin{tabular}{c|cccc|c|cccc|c}
        \toprule
        & \multicolumn{5}{c|}{PACS} & \multicolumn{5}{c}{VLCS} \\
        \cmidrule{2-11} 
        & Art. & Cartoon & Sketch & Photo & Avg. & Caltech & Labelme & Pascal & Sun & Avg. \\
        \midrule
       
        JiGen  & 79.42 & 75.25 & 71.35 & 96.03 & 80.51 & 96.17 & 62.06 & 70.93 & 71.40 & 75.14  \\
        M-ADA  & 64.29 & 72.91 & 67.21 & 88.23 & 73.16 & 74.33 & 48.38 & 45.31 & 33.82 & 50.46 \\
        DG-MMLD  & 81.28 & 77.16 & 72.29 & \underline{96.09} & 81.83 & \textbf{97.01} & 62.20 & 73.01 & \underline{72.49} & \underline{76.18} \\
        D-SAM* \cite{d2018domain} & 77.33 & 72.43 & 77.83 & 95.30 & 80.72 & - & - & - & - & - \\
        Epi-FCR* \cite{li2019episodic} & 82.10 & 77.00 & 73.00 & 93.90 & 81.50 & - & - & - & - & - \\
        FAR* \cite{jin2020feature} & 79.30 & 77.70 & 74.70 & 95.30 & 81.70 & - & - & - & - & - \\
        MetaReg* \cite{NEURIPS2018_647bba34} & \textbf{83.70} & 77.20 & 70.30 & 95.50 & 81.70 & - & - & - & - & - \\
        RSC  & \underline{83.43} & \textbf{80.31} & \textbf{80.85} & 95.99 & \textbf{85.15} & 96.21 & \underline{62.51} & \textbf{73.81} & 72.10 & 76.16 \\
        \midrule
        ResNet-18 & 76.61 & 73.60 & 76.08 & 93.31 & 79.90 & 91.86 & 61.81 & 67.48 & 68.77 & 72.48 \\
        StableNet (ours) & 81.74 & \underline{79.91} & \underline{80.50} & \textbf{96.53} & \underline{84.69} & \underline{96.67} & \textbf{65.36} & \underline{73.59} & \textbf{74.97} & \textbf{77.65}\\
        \bottomrule
    \end{tabular}}
    \label{tab:comp}
\end{table*}

\begin{figure*}[ht]
    \centering
    \begin{subfigure}{0.3\textwidth}
        \centering
        \includegraphics[width=\textwidth]{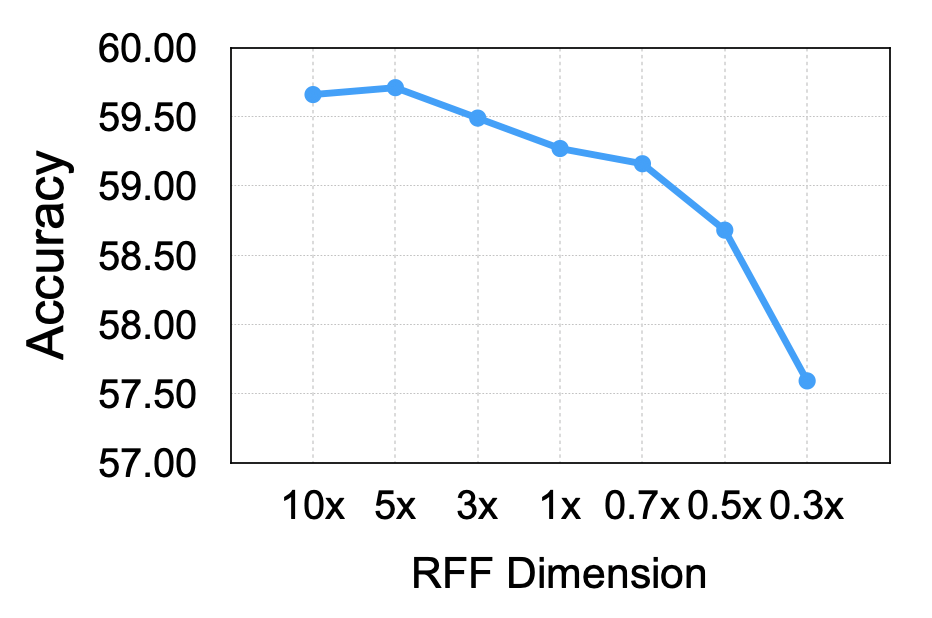}
        \caption{}
    \end{subfigure}
    \begin{subfigure}{0.3\textwidth}
        \centering
        \includegraphics[width=\textwidth]{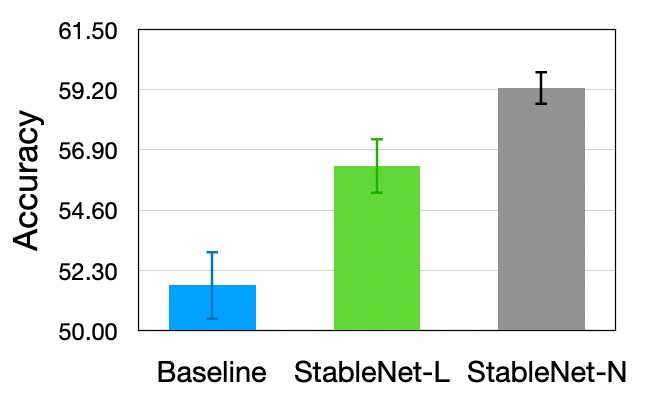}
        \caption{}
    \end{subfigure}
    \begin{subfigure}{0.3\textwidth}
        \centering
        \includegraphics[width=\textwidth]{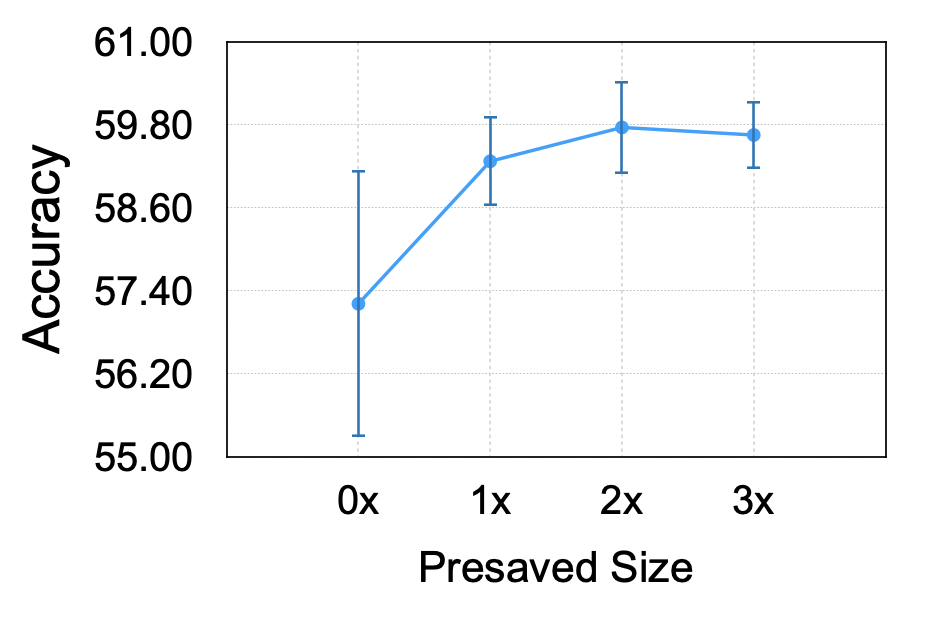}
        \caption{}
    \end{subfigure}
    \vspace{-8pt}
    \caption{Results of ablation study on NICO. All the experiments adopt NICO since NICO consists of a wide range of domains and objects and all domains come from real-world images which make the indication of results more reliable. The RFF dimension in (a) indicates the dimension of Fourier features, where \textsl{10x} indicates that the dimension of Fourier features are 10 times the size of original features and \textsl{0.3x} indicates the sampling ratio is 30\%. \textsl{StableNet-N}  and \textsl{StableNet-L} indicate the original StableNet and the degenerated version of StableNet that only eliminates the linear correlation between features. \textsl{Presaved size} in (c) indicates the dimension of the presaved features and \textsl{0x} indicates no features are saved.} 
    \label{fig:ablation}
    \vspace{-8pt}
\end{figure*}

\subsection{Experimental settings and datasets}
We validate StableNet in a variety of settings. 
To cover more general and challenging cases of distribution shifts, we adopt four experimental settings as follows:  

\noindent \textbf{Unbalanced}. In the common DG setting, the capacities of source domains are assumed to be comparable. However, considering most datasets are a mixture of latent unknown domains, one can hardly assume that the amount of samples from these domains are consistent since these datasets are not generated by equally sampling from latent domains. 
We simulate this scenario with this setting. Domains are split into source domains and target domains. The capacities of various domains can vary significantly. Note that this setting, where the capacities of available domains are unbalanced while the proportion of each class remains consistent across domains, is completely different from the settings of the class imbalance problem. This setting is to evaluate the generalization ability of models when the heterogeneity is unclear and insignificant. 

\noindent \textbf{Flexible}. We consider a more challenging but common in real-world setting where domains for different categories can be various. For instance, birds can be on trees but hardly in the water while fishes are the opposite. If we consider the backgrounds in images as an indicator of domain division, images for class `bird' can be divided into domain `on tree' but cannot into domain `in water' while images for class `fish' are otherwise, resulting in the diversity of domains among different classes. Thus this setting simulates a widely existing scenario in the real-world. In such cases, the level of the distribution shifts varies in different classes, requiring a strong ability of generalization given the statistical correlations between relevant features and category-irrelevant features vary.

\noindent \textbf{Adversarial}. We consider the most challenging scenario, where the model is under adversarial attack and the spurious correlations between domains and labels are strong and misleading. For instance, we assume a scenario where the category `dog' is usually associated with the domain `grass' and the category `cat' with the domain `sofa' in the training data, while the category `dog' is usually associated with the domain `sofa' and the category `cat' with the domain `grass' in the testing data. If the ratio of domain `grass' in the images from class `dog' is significantly higher than others, the predictive model may tend to recognize grass as a dog. 

\noindent \textbf{Classic}. This setting is the same as the common setting in DG. 
The capacities of various domains are comparable. Therefore this setting is to evaluate the generalization ability of models when the heterogeneity of training data is significant and clear, which is less challenging compared with the previous three settings. 




\noindent \textbf{Datasets}. We consider four datasets to carry through these four settings, namely PACS \cite{li2017deeper}, VLCS \cite{torralba2011unbiased}, MNIST-M \cite{ganin2015unsupervised} and NICO \cite{he2020towards}. Introduction to these datasets and details of implementation are in Appendix C.1.

\subsection{Unbalanced setting}

Given this setting requires all the classes in the dataset share the same candidate set of domains, which is incompatible with NICO, we adopt PACS and VLCS for this setting. 
Three domains are considered as source domains and the other one as target. 
To make the amount of data from heterogeneous sources clearly differentiated, we set one domain as the dominant domain.
For each target domain, we randomly select one domain from the source domains as the dominant source domain and adjust the ratio of data from the dominant domain and the other two domains. Details of ratios and partition are shown in Appendix C.2.

Here we show the results when the capacity ratio of three source domains is 5:1:1 in Table \ref{tab:comp_dom} and our method outperforms other methods in all the target domains on both PACS and VLCS. Moreover, StableNet achieves best performance consistently under all the other ratios as shown in Appendix C.2.
These results indicate that the subtle statistical correlations between relevant and irrelevant features are strong enough to significantly harm the generalization across domains. When the correlations are eliminated, the model is able to learn the true connections between relevant features and labels and inference according to them only, thus generalize better. 
For adversarially trained methods like DG-MMLD \cite{matsuura2020domain}, the supervision from minor domains is insufficient and the ability of the model to discriminate irrelevant features is impaired. For augmentation of source domains based methods like M-ADA \cite{qiao2020learning}, the impact of the dominant domain is not diminished while the minor ones are still insignificant after the augmentation. Methods like RSC \cite{huang2020self} adopt regularization to prevent the model from overfitting on source domains and the samples from minor domains can be considered as outliers and ignored.
Therefore, the subtle correlations between relevant features and irrelevant features especially in minor domains are not eliminated.   

\subsection{Unbalanced + flexible setting}

We adopt PACS, VLCS and NICO to evaluate the \textsl{unbalanced + flexible} setting. For PACS and VLCS, we randomly select one domain as the dominant domain for each class, and another domain as the target. 
For NICO, there are 10 domains for each class, 8 out of which are selected as the source and 2 as the target.  We adjust the ratio of the dominant domain to minor domains to adjust the level of distribution shifts. Here we report the results when the dominant ratio is 5:1:1. 
Details and more results of other divisions are shown in Appendix C.3.

The results are shown in Table \ref{tab:comp_dom_flexible}. 
M-ADA and DG-MMLD fail to outperform ResNet-18 on NICO under this setting. M-ADA, which generates images for training with an autoencoder, may fail when the training data are large-scale real-world images and the distribution shifts are not caused by random disturbance. DG-MMLD generates domain labels with clustering and may fail when the data lack explicit heterogeneity or the number of latent domains is too large for clustering. 
In contrast, StableNet shows a strong ability of generalization when the input data are with complicated structure especially real-world images from unlimited resources. StableNet can capture various forms of dependencies and balance the distribution of input data. 
On PACS and VLCS, StableNet also outperforms state-of-the-art methods, showing the effectiveness of removing statistical dependencies between features especially when the source domains for different categories are not consistent. More experimental results are in Appendix C.3.   

\subsection{Unbalanced + flexible + adversarial setting}

To exploit the effect of various levels of adversarial attack, we adopt MNIST-M to evaluate our method owing to the numerous (200) optional domains in MNIST-M. Domains in PACS and VLCS are insufficient to generate multiple adversarial levels. Hence, we generate a new MNIST-M dataset with three rules: 1) for a given category, there is no overlap between the domains in training and testing; 2) a background image is randomly chosen for each category in the training set, and contexts cropped in the same image are assigned as dominant contexts (domains) for another category in test data so that there are strong spurious correlations between labels and domains; 3) the ratio of dominant context to other contexts varies from 9.5:1 to 1:1 to generate settings with different levels of distribution shifts. 
Detailed data generating method, adopted backbone network and sample images are in Appendix C.4.

The results are shown in Table \ref{tab:mnist-m}. As the dominant ratio increases, the spurious correlation between domains and categories becomes stronger so that the performance of predictive models drops. When the imbalance in visual features is significant, our method achieves noticeable improvement compared with baseline methods. For regularization-based methods such as RSC, they tend to weaken the supervision from minor domains which may be considered as outliers and therefore the spurious correlations between irrelevant features and labels are strengthened under adversarial attacks, resulting in even poorer results compared with the vanilla ResNet model. As shown in Table \ref{tab:mnist-m}, RSC fails to outperform vanilla CNNs.


\begin{figure*}[ht]
    \centering
    \includegraphics[width=0.78\textwidth]{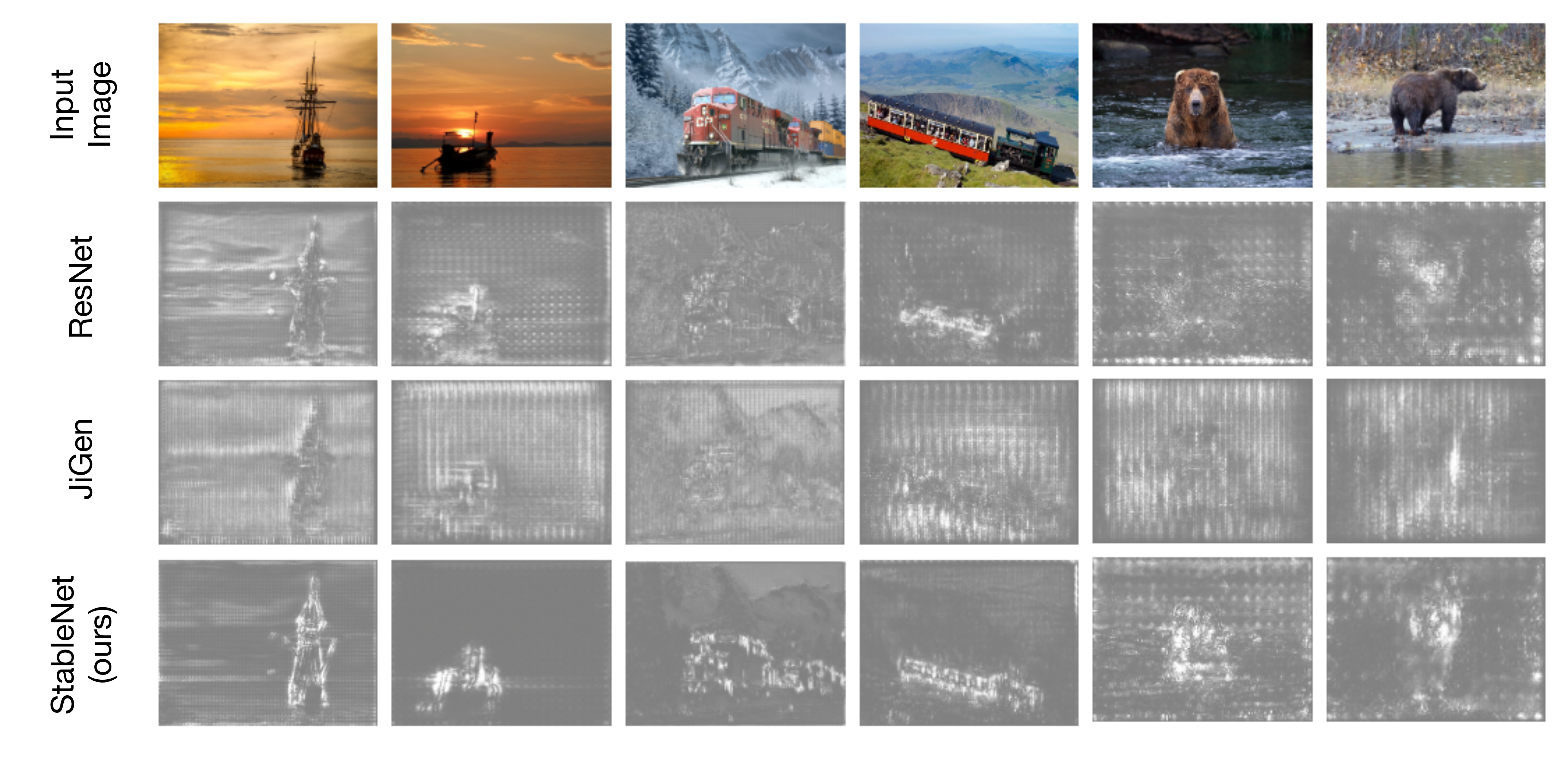}
    \vspace{-8pt}
    \caption{Saliency maps of the ResNet-18 model and StableNet. The brighter the pixel is, the more contributions it makes to prediction.}
    \label{fig:saliency}
    \vspace{-8pt}
\end{figure*}

\subsection{Classic setting}
The \textsl{classic} setting is the same as the common setting in DG. 
Domains are split into source domains and target domains. The capacities of various domains are comparable.
Given this setting requires all the classes in the dataset to share the same candidate set of domains, which is incompatible with NICO, we adopt PACS and VLCS for this setting. We follow the experimental protocol of \cite{carlucci2019domain, matsuura2020domain} for both the datasets and utilize three domains as source domains and the remaining one as the target. 

The results are shown in Table \ref{tab:comp}. On VLCS, StableNet outperforms other state-of-the-art methods in two out of four target cases and achieves the highest average accuracy. On PACS, StableNet achieves the highest accuracy on the target domain `photo' and comparable average accuracy (0.46\% less) compared with the state-of-the-art method, RSC. The accuracy gap between StableNet and baseline indicates that even when the numbers of samples from different source domains are approximately the same, the subtle statistical correlations between relevant features and irrelevant features still hold strong and the model generalizes across domains better when the correlations are eliminated.

\subsection{Ablation study}

StableNet relies on Random Fourier Features sampled from Gaussian to balance the training data. 
The more features are sampled, the more independent the final representations are. In practice, however,  generating more features requires more computational cost. In this ablation study, we exploit the effect of sampling size for Random Fourier Features.
Moreover, inspired by \cite{tanskanen2018random}, one can further reduce the feature dimension by randomly selecting features used to calculate dependence with different ratios. 
Figure \ref{fig:ablation} shows the results of StableNet with different dimensions of Random Fourier Features.  
If we remove all the Random Fourier Features, our regularizer in Equation \ref{eq:learning_weight} degenerates and can only model the linear correlation between features. Figure 2(a) demonstrates the effectiveness of eliminating non-linear dependence between representations. From Figure 2(b), the non-linear dependence is common in vision features and keep deep models from learning true dependence between input images and category labels. 

We further exploit the effect of the size of presaved features and weights in Equation \ref{eq:concat_w} and the results are shown in Figure 2(c). When the size of presaved features is reduced to 0, sample weights are learned inside of each batch, yielding noticeable variance.
Generally, as the presaving size increases, the accuracy raises slightly and the variance drops significantly, indicating that presaved features help to learn sample weights globally and therefore the generalization ability of the model is more stable.






\subsection{Saliency map}
An intuitive type of explanation for image classification models is to identify pixels that have a strong influence on the final decision \cite{smilkov2017smoothgrad}. To demonstrate whether the model focuses on the object or the context (domain) while conducting prediction, we visualize the gradient of the class score function with respect to the input pixels. In the case of stable learning, we adopt the same backbone architecture for all methods, so that we adopt \textsl{smoothed} gradient as suggested by \cite{adebayo2018sanity}, which generates saliency maps depending on the learned parameters of the models instead of the architecture. 
Visualization results are shown in Figure \ref{fig:saliency}. Saliency maps of the baseline model show that various contexts draw noticeable focus of the classifier while fail to make decisive contributions to our model. More visualization results are in Appendix C.6, which further demonstrate that StableNet focuses more on visual parts which are both distinguishing and invariant when the postures or positions of objects vary.


\section{Conclusion}
In the paper, to improve the generalization of deep models under distribution shifts, we proposed a novel method called StableNet which can eliminate the statistical correlation between relevant and irrelevant features via sample weighting. Extensive experiments across a wide range of settings demonstrated the effectiveness of our method.  


\section*{Acknowledgement}
This work was supported in part by National Key R\&D Program of China (No. 2018AAA0102004, No. 2020AAA0106300), National Natural Science Foundation of China (No. U1936219, 61521002, 61772304), Beijing Academy of Artificial Intelligence (BAAI), and a grant from the Institute for Guo Qiang, Tsinghua University.

\balance
{\small
\bibliographystyle{ieee_fullname}
\bibliography{cvpr}
}

\clearpage

\appendix
\section{Method}
\renewcommand{\algorithmicrequire}{\textbf{Input:}}
\renewcommand{\algorithmicensure}{\textbf{Output:}}
\begin{algorithm*}[ht]
    \caption{\emph{Training procedure of \setlength{\parindent}{4em}StableNet}}
    \label{alg}
    \begin{algorithmic}[1]
        \REQUIRE EPOCH\_NUMBER, BALANCING\_EPOCH\_NUMBER
        \ENSURE Learned model
        \FOR{epoch $\leftarrow$ 1 to EPOCH\_NUMBER}
            \FOR{batch $\leftarrow$ 1 to BATCH\_NUMBER}
                \STATE Forward propagate
                \STATE Reload global features as Equation \ref{eq:concat_z}
                \FOR{epoch\_balancing $\leftarrow$ 1 to BALANCING\_EPOCH\_NUMBER}
                    \STATE Optimize sample weights as Equation \ref{eq:learning_weight}
                \ENDFOR
                \STATE Back propagate with weighted prediction loss as Equation \ref{eq:weighted_loss}
                \STATE Save features and weights as Equation \ref{eq:save_w}
            \ENDFOR
        \ENDFOR
    \end{algorithmic}
\end{algorithm*}

\subsection{Detailed Training Procedure of StableNet} \label{app:B_1}
In the training phase, StableNet learns a set of sample weights for each batch with the global knowledge of correlations between features saved before. The parameters of the model and the sample weights are optimized iteratively. 

As shown in Algorithm \ref{alg}, for each input batch $(\mathbf{X}_L, \mathbf{Y}_L)$, the corresponding representations $\mathbf{Z}_L = f(\mathbf{X}_L)$ is concatenated with pre-saved global representations $\mathbf{Z}_{G1}, \mathbf{Z}_{G2}, \cdots, \mathbf{Z}_{Gk}$, as shown in Equation \ref{eq:concat_z}.
\begin{equation}\label{eq:concat_z}
    \mathbf{Z}_{O} = \text{Concat}\left(\mathbf{Z}_{G1}, \mathbf{Z}_{G2},\cdots, \mathbf{Z}_{Gk}, \mathbf{Z}_{L} \right).
\end{equation}
Then the model learns the local weights with global weights $\mathbf{w}_{G1}, \mathbf{w}_{G2}, \cdots, \mathbf{w}_{Gk}$ by optimizing Equation \ref{eq:learning_weight}.
\begin{equation}\label{eq:learning_weight}
    \mathbf{w}_L = \underset{\mathbf{w} \in \Delta_B}{\arg \min} \sum_{1 \le i < j \le m_Z} \left\Vert\hat{\Sigma}_{\mathbf{Z}_{O_{:,i}}\mathbf{Z}_{O_{:, j}};\mathbf{w}_O}\right\Vert_F^2,
\end{equation}
where $B$ is the batch size and
\begin{equation*}\label{eq:w_o}
    \mathbf{w}_O = \text{Concat}\left(\mathbf{w}_{G1}, \mathbf{w}_{G2},\cdots, \mathbf{w}_{Gk}, \mathbf{w} \right).
\end{equation*}
The loss for optimizing the representation extractor $f$ and the classifier $g$ is calculated by conducting sample weights $\mathbf{w}_L$ and penalties for input samples as shown in Equation \ref{eq:weighted_loss}.
\begin{equation}\label{eq:weighted_loss}
    L_{f,g} = \sum_{i=1}^{B} \mathbf{w}_{L_i} L(g(f(\mathbf{X}_{L_i})), \mathbf{Y}_{L_i}).
\end{equation}
The present features and weights are integrated with the previous global features and weights as shown in Equation \ref{eq:save_w}.
\begin{equation}\label{eq:save_w}
    \begin{aligned}
        \mathbf{Z}_{Gi}' & = \alpha_{i}\mathbf{Z}_{Gi} + (1 - \alpha_i)\mathbf{Z}_{L}, \\
        \quad \quad \mathbf{w}_{Gi}' & = \alpha_{i}\mathbf{Z}_{Gi} + (1 - \alpha_i)\mathbf{w}_{L}.
    \end{aligned}
\end{equation}

In the inference phase, given the back propagation is disabled, StableNet escapes the sample weighting phase and conduct prediction directly.

In practice, the optimization also requires a regularizer of  weight decay. We set the weight of the regularizer to 0.3 and the learning rate for sample weights to 3.0 unless otherwise noted. 

\subsection{Other Approaches for Independence Test} \label{app:B_2}
Despite the fact that there are several approaches to be used as the supervision of feature independence, they can hardly be used to optimize deep models within acceptable cost. For instance, Hilbert-Schmidt Independence Criterion (HSIC), which can be used in independent component analysis (ICA) \cite{gretton2005measuring}, is applied as a criterion for feature decorrelation in \cite{bahng2019learning}. However, the calculation of HSIC requires noticeable computational cost which grows as the batch size of training data increases, so it is inapplicable to training deep models on large datasets. Another approach, Mutual Information (MI), requires variational bounds to estimate \cite{poole2019variational}, which are hard to be assembled with common convolutional networks such as ResNet \cite{he2016deep}.

\subsection{Correlation between the Unbalanced Setting and the Problem of Unbalanced Classes}
There are many works focus on the class imbalance problem\cite{li2020overcoming, lin2017focal, wang2020deep}. But our \textsl{unbalanced} setting is a completely different problem. The key of problem of class imbalance is learning a better classifier to improve the recognition accuracy of minor classes, while the \textsl{unbalanced} setting valuates the ability of learning the correlation between object-relevant features and labels 
using the heterogeneity of unbalanced domains. In our setting, the proportion of different classes , which our settings do not focus on, may be different in both the \textsl{unbalanced} setting and the \textsl{classic} setting. Experiments show that methods for the class imbalance problem is not effective (comparable or worse compared with ResNet-18) in the \textsl{unbalanced} setting.

\subsection{Correlation between our method and feature disentanglement}
There are many works attempt learning disentangled features for robust or explainable representations\cite{huang2020unsupervised, burgess2018understanding, hsieh2018learning} with strong constrains on features. These disentanglement methods such as VAE \cite{kingma2013auto} force features to be disentangled, which changes the semantic implication of features, while StableNet learns sample weights to adjust the data structure while the semantic of features is not effected given disentanglement is not our target but a path to learn true correlation between relevant features and labels. We fail to train VAE-based models that outperform the ResNet-18 model in all of our settings.

\section{Experiments}

\subsection{Datasets}
We adopt 4 datasets to conduct experiments in our 4 settings. We briefly introduce them as follows.

\textbf{MNIST-M} is generated by the method in the original paper, which is blending digits from the original MNIST dataset over patches extracted from images in BSDS500 dataset.

\textbf{VLCS} consists of 5 object categories shared by the PASCAL VOC 2007, LabelMe, Caltech and Sun datasets. We follow the standard protocol of \cite{ghifary2015domain} and divide each domain into a training set (70\%) and validation set (30\%) randomly. 

\textbf{PACS} is a widely used benchmark for domain generalization which consists of 7 object categories spanning 4 image styles, namely \textsl{photo, art-painting, cartoon and sketch}. We adopt the protocol in \cite{li2017deeper} to split the training and val set.

\textbf{NICO} is dedicately designed for Non-I.I.D (distribution shifts) image classification. The images from each category can be wildly various and labeled with 10 contexts.

\subsection{Training Details}
\begin{table}[ht]
    \centering
    \caption{The structure of CNNs for MNIST-M}
    \scalebox{0.85}{
    \begin{tabular}{c|c}
        \toprule
        Layer & Details \\
        \midrule
        Input & $3 \times 28 \times 28$ \\
        Conv & Kernel Size 7, Stride 1, Out Channel 32, BN, ReLU \\
        Conv & Kernel Size 5, Stride 2, Out Channel 32, BN, ReLU \\
        Dropout & $p=0.4$\\
        Conv & Kernel Size 3, Stride 1, Out Channel 64, BN, ReLU \\
        Conv & Kernel Size 3, Stride 2, Out Channel 64, BN, ReLU \\
        Dropout & $p=0.4$\\
        FC & Out Channel 16, ReLU\\
        SoftMax & Class\_Num\\
        \bottomrule
    \end{tabular}}
    \label{tab:Shallow}
\end{table}
For PACS and VLCS, we adopt ResNet-18\cite{he2016deep} as the backbone model. Given the images from PACS and VLCS are not sufficient to train a randomly initialized deep model like ResNet-18, we use the weights pretrained on ImageNet\cite{deng2009imagenet}. For NICO, randomly initialized ResNet-18 is used as the backbone model. For PACS and VLCS, We train all the methods 30 epochs. The initial learning rate is 0.01 and decays with a rate of 0.1 at epoch 24. For NICO, we train all the methods 60 epochs. We set initial learning rate to 0.02 which decays with a rate of 0.1 at epoch 30. The weight decay is set to 0.0005 for all the three datasets. For MNIST-M, we use a 4-layer convolutional network as the backbone and the structure is shown in Table \ref{tab:Shallow}. The models are trained 30 epochs. The learning rate is 0.02 and decays with a rate of 0.1 at epoch 20. The weight decay is 0.001.

For JiGen \cite{carlucci2019domain}, DG-MMLD\cite{matsuura2020domain} and RSC \cite{huang2020self}, we adopt the code published by the authors of original papers and the hyperparameters reported in original papers. For M-ADA \cite{qiao2020learning}, we train the model for 5000 iterations on MNIST-M, PACS and VLCS, 8000 iterations on NICO.

\begin{table*}[ht]
    \centering
    \caption{Results of the \textsl{unbalanced} setting when the domain ratio is 3:1:1 on PACS and VLCS. The reported results are average over three repetitions of each run. The title of each column indicates the name of the domain used as target. The best results of all methods are highlighted with the bold font.}
    \scalebox{0.85}{
    \begin{tabular}{c|cccc|c|cccc|c}
        \toprule
         & \multicolumn{5}{c|}{PACS} & \multicolumn{5}{c}{VLCS} \\
         \cmidrule{2-11} 
         & Art. & Cartoon & Sketch & Photo & Avg. & Caltech & Labelme & Pascal & Sun & Avg. \\
        \midrule
        
        JiGen \cite{carlucci2019domain} & 76.25 & 71.60 & 68.79 & 92.05 & 77.17 & 84.82 & 58.45 & 64.42 & 56.06 & 65.94 \\
        M-ADA \cite{qiao2020learning} & 65.76 & 66.98 & 54.49 & 90.33 & 69.39 & 66.25 & 53.13 & 43.80 & 47.30 & 52.62\\
        DG-MMLD \cite{matsuura2020domain} & 81.98 & 73.53 & 74.32 & 92.65 & 80.61 & 79.42 & 58.29 & 64.96 & 51.16 & 63.46 \\
        Focal loss \cite{lin2017focal} & 75.56 & 70.13 & 67.91 & 90.75 & 76.09 & 75.22 & 53.98 & 56.61 & 53.88 & 59.92\\
        RSC \cite{huang2020self} & \textbf{84.44} & 74.82 & 70.87 & 94.66 & 81.20 & 82.31 & 59.55 & \textbf{65.32} & 56.11 & 65.82 \\
        \midrule
        ResNet-18 & 75.90 & 73.34 & 69.22 & 93.82 & 78.07 & 81.66 & 61.75 & 60.20 & 57.33 & 65.24\\
        StableNet (ours) & 79.15  & \textbf{79.96} & \textbf{75.44} & \textbf{95.19} & \textbf{82.44} & \textbf{85.51}  & \textbf{65.46} & 63.65 & \textbf{59.93} & \textbf{68.64} \\
        \bottomrule
    \end{tabular}}
    \label{tab:dom3}
\end{table*}

        
\begin{table*}[ht]
    \centering
    \caption{Results of the \textsl{unbalanced} setting when the domain ratio is 2:1:1 on PACS and VLCS. For the number of each run and the use of font, see Table \ref{tab:dom3}.}
    \scalebox{0.85}{
    \begin{tabular}{c|cccc|c|cccc|c}
        \toprule
         & \multicolumn{5}{c|}{PACS} & \multicolumn{5}{c}{VLCS} \\
         \cmidrule{2-11} 
         & Art. & Cartoon & Sketch & Photo & Avg. & Caltech & Labelme & Pascal & Sun & Avg. \\
        \midrule
        
        JiGen & 74.20 & 71.60 & 68.88 & 93.15 & 76.96 & 89.29 & 64.41 & 66.38 & 55.46 & 68.89 \\
        M-ADA & 74.32 & 57.03 & 59.40 & 93.67 & 71.11 & 87.03 & 63.52 & 54.58 & 52.03 & 64.29 \\
        DG-MMLD & 79.94 & 71.67 & 68.58 & 92.85 & 78.26 & 86.39 & 66.67 & 65.39 & 54.59 & 68.26 \\
        RSC & 80.51 & 69.53 & 68.96 & \textbf{94.36} & 78.34 & 87.59 & 65.05 & 65.67 & 54.37 & 68.17\\
        \midrule
        ResNet-18 & 75.75 & 68.35 & 64.59 & 93.37 & 75.52 & 84.77 & 64.62 & 65.38 & 52.59 & 66.84\\
        StableNet (ours) & \textbf{80.56}  & \textbf{75.33} & \textbf{71.48} & 94.29 & \textbf{80.42} & \textbf{91.29}  & \textbf{67.87} & \textbf{66.72} & \textbf{56.55} & \textbf{70.61} \\
        \bottomrule
    \end{tabular}}
    \label{tab:dom2}
\end{table*}


\begin{table*}[ht]
    \centering
    \caption{Results of the \textsl{unbalanced + flexible} setting when the domain ratio is 3:1:1 on PACS and VLCS. For the number of each run and the use of font, see Table \ref{tab:dom3}.}
    \scalebox{0.85}{
    \begin{tabular}{c|cccc|c|cccc|c}
        \toprule
         & \multicolumn{5}{c|}{PACS} & \multicolumn{5}{c}{VLCS} \\
         \cmidrule{2-11} 
         & Art. & Cartoon & Sketch & Photo & Overall & Caltech & Labelme & Pascal & Sun & Overall \\
        \midrule
        
        JiGen  & 57.95 & - & 27.70 & 90.02 & 44.03 & 97.71 & - & 61.50 & 67.12 & 76.31 \\
        M-ADA  & 49.58 & - & 15.21 & 79.92 & 33.33 & 97.55 & - & 44.17 & 60.32 & 69.09\\
        DG-MMLD & 68.57 & - & 37.65 & 91.91 & 52.56 & \textbf{99.85} & - & 65.37 & 67.16 & 78.08 \\
        RSC & 66.10 & - & 26.48 & 90.52 & 44.55 & 99.82 & - & 56.79 & 65.25 & 75.09 \\
        \midrule
        ResNet-18 & 57.11 & - & 28.54 & 89.79 & 45.07 & 99.25 & - & 63.37 & 63.55 & 75.93\\
        StableNet (ours) & \textbf{71.82} & - & \textbf{38.16} & \textbf{92.36} &\textbf{54.10} & 99.75 & - & \textbf{67.53} & \textbf{68.89} & \textbf{79.28}\\  
        \bottomrule
    \end{tabular}}
    \label{tab:flex3}
\end{table*}

\begin{table*}[ht]
    \centering
    \caption{Results of the \textsl{unbalanced + flexible} setting when the domain ratio is 2:1:1 on PACS and VLCS. For the number of each run and the use of font, see Table \ref{tab:dom3}.}
    \scalebox{0.85}{
    \begin{tabular}{c|cccc|c|cccc|c}
        \toprule
         & \multicolumn{5}{c|}{PACS} & \multicolumn{5}{c}{VLCS} \\
         \cmidrule{2-11} 
         & Art. & Cartoon & Sketch & Photo & Overall & Caltech & Labelme & Pascal & Sun & Overall \\
        \midrule
        
        JiGen  & 50.14 & 87.08 & 16.45 & 81.55 & 40.12 & 7.11 & 39.55 & 39.85 & 31.90 & 37.99 \\
        M-ADA  & 43.86 & 81.60 & 9.88 & 63.51 & 32.17 & 6.02 & 35.34 & 29.59 & 22.22 & 27.88 \\
        DG-MMLD  & 66.46 & 91.78 & \textbf{30.64} & 80.07 & 51.71 & 10.44 & 44.57 & 41.85 & \textbf{42.86} & 41.72 \\
        RSC  & 53.02 & 88.09 & 16.59 & 83.58 & 41.24 & 1.20 & 36.87 & 26.89 & 30.00 & 30.78 \\
        \midrule
        ResNet-18 & 56.62 & 80.56 & 20.98 & 77.58 & 42.83 & 45.42 & 33.72 & 39.17 & 23.81 & 36.49\\
        StableNet (ours) & \textbf{67.85}  & \textbf{92.01} & 29.57 & \textbf{88.92} & \textbf{52.68} & \textbf{59.88}  & \textbf{48.15} & \textbf{49.58} & \textbf{42.86} & \textbf{49.27} \\
        \bottomrule
    \end{tabular}}
    \label{tab:flex2}
\end{table*}

\begin{table*}[ht]
    \centering
    \caption{Results of the \textsl{unbalanced + flexible} setting when the domain ratio is 1:1:1 on PACS and VLCS. For the number of each run and the use of font, see Table \ref{tab:dom3}.}
    \scalebox{0.85}{
    \begin{tabular}{c|cccc|c|cccc|c}
        \toprule
         & \multicolumn{5}{c|}{PACS} & \multicolumn{5}{c}{VLCS} \\
         \cmidrule{2-11} 
         & Art. & Cartoon & Sketch & Photo & Overall. & Caltech & Labelme & Pascal & Sun & Overall. \\
        \midrule
        
        JiGen  & 61.84 & 94.37 & 26.67 & 87.76 & 44.91 & 99.97 & 17.93 & 59.84 & 53.34 & 69.71 \\
        M-ADA  & 3.06 & 80.45 & 22.87 & 85.38 & 36.42 & 89.92 & 5.75 & 46.30 & 41.54 & 58.02 \\
        DG-MMLD & 65.77 & 97.19 & \textbf{43.44} & 89.87 & 57.07 & \textbf{100.00} & 20.00 & 60.43 & 55.12 & 70.61 \\
        RSC  & 62.64 & 96.97 & 23.99 & 89.75 & 43.69 & \textbf{100.00} & 15.17 & 54.03 & 50.09 & 66.87 \\
        \midrule
        ResNet-18 & 64.68 & 95.56 & 24.39 & 90.25 & 44.15 & 96.25 & 3.91 & 49.34 & 49.30 & 63.93\\
        StableNet (ours) & \textbf{67.66}  & \textbf{98.52} & 41.90 & \textbf{95.85} & \textbf{57.41} & \textbf{100.00}  & \textbf{24.14} & \textbf{63.60} & \textbf{63.19} & \textbf{74.70} \\
        \bottomrule
    \end{tabular}}
    \label{tab:flex1}
\end{table*}

\subsection{More Results and Data Split Details of Unbalanced Setting} \label{app:unbalanced}
In the \textsl{unbalanced} setting, , we randomly choose a dominant domain for each target domain for both PACS and VLCS. The ratio of data amount from dominant domain to other domains are 5:1:1 in Section 4.2. To simulate a more general setting, we also evaluate DG methods on settings where domain ratio is 3:1:1 and 2:1:1. The numbers of samples from each domain on PACS are shown in Table \ref{tab:comp_dom_PACS_5}, \ref{tab:comp_dom_PACS_3}, and \ref{tab:comp_dom_PACS_2}. The numbers of samples from each domain on VLCS are shown in Table \ref{tab:comp_dom_VLCS_5}, \ref{tab:comp_dom_VLCS_3} and \ref{tab:comp_dom_VLCS_2}. 

The results of the \textsl{unbalanced} setting when the domain ratio is 3:1:1 on PACS and VLCS are shown in Table \ref{tab:dom3}. StableNet outperforms all the other state-of-the-art methods on all the target domains on VLCS and three out of four domains on PACS. StableNet achieves the highest average accuracy across four domains both on PACS and VLCS. The results of the \textsl{unbalanced} setting when the domain ratio is 2:1:1 on PACS and VLCS are shown in Table \ref{tab:dom2}. StableNet also shows strong ability of generalization consistently in all the three \textsl{unbalanced} settings, which indicates the effectiveness of StableNet on more general settings of distribution shift problems.

\subsection{More Results and Data Split Details of Unbalanced + Flexible Setting} \label{app:flexible}
In the \textsl{unbalanced + flexible} setting, we randomly choose a dominant domain for each target domain for both PACS and VLCS. The ratio of data amount from dominant domain to other domains are 5:1:1 in Section 4.3. We evaluate DG methods on \textsl{unbalanced + flexible} settings where the domain ratio is 5:1:1, 3:1:1, 2:1:1 and 1:1:1, respectively. Note that when the ratio is set to 1:1:1, this setting degenerates to \textsl{flexible} setting where the numbers of samples from different domains are approximately the same.

Actually, the classic DG setting is not sufficient for evaluating the generalization ability of models since the target domain is unique when the source domains are determined. So given the source domains, only the performance on a given target domain is evaluated while the goal of generalization is to generalize to any target domains. The \textsl{unbalanced + flexible} setting, however, can evaluate the ability of generalization to any domains since the model are tested on all the domains for given source domains. In other words, the model is trained once and tested on all the domains, while in the classic DG setting, the model is tested on one single domain for each training. Moreover, there are no overlap between source domains and the target domain for a single class so the generalization ability is evaluated.

The results of the \textsl{unbalanced + flexible} setting when the domain ratio is 3:1:1 and 2:1:1 on PACS and VLCS are shown in Table \ref{tab:flex3} and Table \ref{tab:flex2}, respectively. Note that the target domain for each class is randomly chosen so that there are two domains containing no test data. Details of data split of this setting are shown in Table \ref{tab:comp_dom_PACS_5} and Table \ref{tab:comp_dom_VLCS_5}. Given the capacities of different target domains can be significantly various, we report the weighted average accuracy (denoted by `overall' in the table) of all domains instead of naive average accuracy for the \textsl{unbalanced + flexible} setting, which is different from all the reported average accuracy for other settings.
We show the accuracy of methods on all the domains. StableNet outperforms all the other state-of-the-art methods on almost all the target domains and on average accuracy on PACS and VLCS. Moreover, StableNet achieves the highest average accuracy across four domains both on PACS and VLCS. The results of the \textsl{unbalanced + flexible} setting when the domain ratio is 1:1:1 on PACS and VLCS are shown in Table \ref{tab:flex1}, where StableNet also outperforms other state-of-the-art counterparts.

\begin{table*}[ht]
    \centering
    \caption{Data split details of \textsl{unbalanced} setting on PACS dataset when the domain ratio is 5:1:1. The dominant domain for each target domain is highlighted with the bold font.}
    \begin{tabular}{lll|l}
        \toprule 
        \multicolumn{3}{c|}{Source} & \multicolumn{1}{c}{Target} \\
        \midrule
        \textbf{Art painting: 2048} & {Cartoon: 405} & {Photo: 405} & {Sketch: 784}  \\
        \textbf{Sketch: 3929} & {Art painting: 779} & {Cartoon: 779} & {Photo: 331}  \\
        \textbf{Photo: 1670} &  {Art painting: 327} & {Sketch: 327} & {Cartoon: 466}  \\
        \textbf{Cartoon: 2344} & {Photo: 463} & {Sketch: 463} & {Art painting: 407} \\
        \bottomrule
    \end{tabular}
    \label{tab:comp_dom_PACS_5}
\end{table*}

\begin{table*}[ht]
    \centering
    \caption{Data split details of \textsl{unbalanced} setting on VLCS dataset when the domain ratio is 5:1:1. The dominant domain for each target domain is highlighted with the bold font.}
    \begin{tabular}{lll|l}
        \toprule
        \multicolumn{3}{c|}{Source} & \multicolumn{1}{c}{Target} \\
        \midrule
        \textbf{Caltech: 991} & {Labelme: 196} & {Pascal: 196} & {Sun: 458}  \\
        \textbf{Sun: 2297} & {Caltech: 350} & {Labelme: 372} & {Pascal: 470}  \\
        \textbf{Pascal: 2363} & {Caltech: 448} & {Sun: 401} & {Labelme: 370} \\
        \textbf{Labelme:1589} & {Pascal: 367} &  {Sun: 367} & {Caltech: 196}  \\
        \bottomrule 
    \end{tabular}
    \label{tab:comp_dom_VLCS_5}
\end{table*}

\begin{table*}[ht]
    \centering
    \caption{Data split details of \textsl{unbalanced} setting on PACS dataset when the domain ratio is 3:1:1. The dominant domain for each target domain is highlighted with the bold font.}
    \begin{tabular}{lll|l}
        \toprule 
        \multicolumn{3}{c|}{Source} & \multicolumn{1}{c}{Target} \\
        \midrule
        \textbf{Art painting: 2048} & {Cartoon: 678} & {Photo: 678} & {Sketch: 784}  \\
        \textbf{Sketch: 3929} & {Art painting: 1217} & {Cartoon: 1238} & {Photo: 331}  \\
        \textbf{Photo: 1670} &  {Art painting: 551} & {Sketch: 538} & {Cartoon: 466}  \\
        \textbf{Cartoon: 2344} & {Photo: 776} & {Sketch: 761} & {Art painting: 407} \\
        \bottomrule
    \end{tabular}
    \label{tab:comp_dom_PACS_3}
\end{table*}

\begin{table*}[ht]
    \centering
    \caption{Data split details of \textsl{unbalanced} setting on VLCS dataset when the domain ratio is 3:1:1. The dominant domain for each target domain is highlighted with the bold font.}
    \begin{tabular}{lll|l}
        \toprule
        \multicolumn{3}{c|}{Source} & \multicolumn{1}{c}{Target} \\
        \midrule
        \textbf{Caltech: 991} & {Labelme: 327} & {Pascal: 327} & {Sun: 458}  \\
        \textbf{Sun: 2297} & {Caltech: 473} & {Labelme: 583} & {Pascal: 470}  \\
        \textbf{Pascal: 2363} & {Caltech: 641} & {Sun: 647} & {Labelme: 370} \\
        \textbf{Labelme: 1859} & {Pascal: 616} &  {Sun: 612} & {Caltech: 196}  \\
        \bottomrule 
    \end{tabular}
    \label{tab:comp_dom_VLCS_3}
\end{table*}

\begin{table*}[ht]
    \centering
    \caption{Data split details of \textsl{unbalanced} setting on PACS dataset when the domain ratio is 2:1:1. The dominant domain for each target domain is highlighted with the bold font.}
    \begin{tabular}{lll|l}
        \toprule 
        \multicolumn{3}{c|}{Source} & \multicolumn{1}{c}{Target} \\
        \midrule
        \textbf{Art painting: 2048} & {Cartoon: 1020} & {Photo: 1020} & {Sketch: 784}  \\
        \textbf{Sketch: 3929} & {Art painting: 1424} & {Cartoon: 1681} & {Photo: 331}  \\
        \textbf{Photo: 1670} &  {Art painting: 831} & {Sketch: 715} & {Cartoon: 466}  \\
        \textbf{Cartoon: 2344} & {Photo: 1136} & {Sketch: 1062} & {Art painting: 407} \\
        \bottomrule
    \end{tabular}
    \label{tab:comp_dom_PACS_2}
\end{table*}

\begin{table*}[ht]
    \centering
    \caption{Data split details of \textsl{unbalanced} setting on VLCS dataset when the domain ratio is 2:1:1. The dominant domain for each target domain is highlighted with the bold font.}
    \begin{tabular}{lll|l}
        \toprule
        \multicolumn{3}{c|}{Source} & \multicolumn{1}{c}{Target} \\
        \midrule
        \textbf{Caltech: 991} & {Labelme: 467} & {Pascal: 494} & {Sun: 458}  \\
        \textbf{Sun: 2297} & {Caltech: 627} & {Labelme: 846} & {Pascal: 470}  \\
        \textbf{Pascal: 2363} & {Caltech: 855} & {Sun: 953} & {Labelme: 370} \\
        \textbf{Labelme: 1859} & {Pascal: 927} &  {Sun: 914} & {Caltech: 196}  \\
        \bottomrule 
    \end{tabular}
    \label{tab:comp_dom_VLCS_2}
\end{table*}

\begin{table*}[ht]
    \centering
    \caption{Data split details of \textsl{unbalanced + flexible} setting on PACS dataset when the domain ratio is 5:1:1. The dominant domain for each target domain is highlighted with the bold font.}
    \begin{tabular}{l|lll|l}
        \toprule
        {Class} & \multicolumn{3}{c|}{Source} & \multicolumn{1}{c}{Target} \\
        \midrule
        {Dog} & \textbf{Cartoon: 350} & {Art painting: 70} &  {Photo: 70} & {Sketch: 772}  \\
        {Elephant} & \textbf{Cartoon: 411} & {Art painting: 82} & {Sketch: 82}&  {Photo: 202}  \\
        {Giraffe} & \textbf{Photo: 163} & {Art painting: 32} & {Cartoon: 32} &  {Sketch: 753}  \\
        {Guitar} & \textbf{Photo: 167}  & {Cartoon: 33}& {Sketch: 33} & {Art painting: 184} \\
        {Horse} & \textbf{Cartoon: 291} & {Photo: 58} & {Sketch: 58} & {Art painting: 201} \\
        {House} & \textbf{Cartoon: 259} &  {Art painting: 51} & {Sketch: 51} & {Photo: 280} \\
        {Person} & \textbf{Art painting: 404} & {Cartoon: 80} & {Photo: 80} & {Sketch: 160} \\
        \bottomrule 
    \end{tabular}
    \label{tab:comp_dom_PACS_5}
\end{table*}

\begin{table*}
    \centering
    \caption{Data split details of \textsl{unbalanced + flexible} setting on VLCS dataset when the domain ratio is 5:1:1. The dominant domain for each target domain is highlighted with the bold font.}
    \begin {tabular}{l|lll|l}
        \toprule
        {Class} & \multicolumn {3}{c|}{Source} & \multicolumn{1}{c}{Target} \\
        \midrule
        {0} & \textbf{Labelme: 39} & {Caltech: 7} &  {Pascal: 7} & {Sun: 14}  \\
        {1} & \textbf{Labelme: 592} & {Caltech: 86} & {Sun: 118}&  {Pascal: 489}  \\
        {2} & \textbf{Pascal: 210} & {Caltech: 42} & {Labelme: 42} &  {Sun: 725}  \\
        {3} & \textbf{Pascal: 205}  & {Labelme: 29}& {Sun: 21} & {Caltech: 47} \\
        {4} & \textbf{Labelme: 606} & {Pascal: 121} & {Sun: 121} & {Caltech: 609} \\
        \bottomrule
    \end{tabular}
    \label{tab:comp_dom_VLCS_5}
\end{table*}

\begin{table*}[ht]
    \centering
    \caption{Data split details of \textsl{unbalanced + flexible} setting on PACS dataset when the domain ratio is 3:1:1. The dominant domain for each target domain is highlighted with the bold font.}
    \begin{tabular}{l|lll|l}
        \toprule
        {Class} & \multicolumn{3}{c|}{Source} & \multicolumn{1}{c}{Target} \\
        \midrule
        {Dog} & \textbf{Cartoon: 389} & {Art painting: 129} &  {Photo: 129} & {Sketch: 772}  \\
        {Elephant} & \textbf{Cartoon: 457} & {Art painting: 152} & {Sketch: 152}&  {Photo: 202}  \\
        {Giraffe} & \textbf{Photo: 182} & {Art painting: 60} & {Cartoon: 60} &  {Sketch: 753}  \\
        {Guitar} & \textbf{Photo: 186}  & {Cartoon: 61}& {Sketch: 61} & {Art painting: 184} \\
        {Horse} & \textbf{Cartoon: 324} & {Photo: 108} & {Sketch: 108} & {Art painting: 201} \\
        {House} & \textbf{Cartoon: 288} &  {Art painting: 95} & {Sketch: 80} & {Photo: 280} \\
        {Person} & \textbf{Art painting: 449} & {Cartoon: 149} & {Photo: 149} & {Sketch: 160} \\
        \bottomrule 
    \end{tabular}
\end{table*}

\begin{table*}
    \centering
    \caption{Data split details of \textsl{unbalanced + flexible} setting on VLCS dataset when the domain ratio is 3:1:1. The dominant domain for each target domain is highlighted with the bold font.}
    \begin {tabular}{l|lll|l}
        \toprule
        {Class} & \multicolumn {3}{c|}{Source} & \multicolumn{1}{c}{Target} \\
        \midrule
        {0} & \textbf{Labelme: 56} & {Caltech: 18} &  {Pascal: 18} & {Sun: 14}  \\
        {1} & \textbf{Labelme: 846} & {Caltech: 86} & {Sun: 281}&  {Pascal: 489}  \\
        {2} & \textbf{Pascal: 300} & {Caltech: 83} & {Labelme: 62} &  {Sun: 725}  \\
        {3} & \textbf{Pascal: 294}  & {Labelme: 29}& {Sun: 21} & {Caltech: 47} \\
        {4} & \textbf{Labelme: 866} & {Pascal: 288} & {Sun: 288} & {Caltech: 609} \\
        \bottomrule
    \end{tabular}
\end{table*}

\begin{table*}[ht]
    \centering
    \caption{Data split details of \textsl{unbalanced + flexible} setting on PACS dataset when the domain ratio is 2:1:1. The dominant domain for each target domain is highlighted with the bold font.}
    \begin{tabular}{l|lll|l}
        \toprule
        {Class} & \multicolumn{3}{c|}{Source} & \multicolumn{1}{c}{Target} \\
        \midrule
        {Dog} & \textbf{Photo: 189} & {Art painting: 94} &  {Cartoon: 94} & {Sketch: 772}  \\
        {Elephant} & \textbf{Art painting: 255} & {Sketch: 127} & {Cartoon: 127}&  {Photo: 202}  \\
        {Giraffe} & \textbf{Photo: 182} & {Art painting: 90} & {Cartoon: 90} &  {Sketch: 753}  \\
        {Guitar} & \textbf{Cartoon: 135}  & {Art painting: 67}& {Sketch: 67} & {Photo: 186} \\
        {Horse} & \textbf{Sketch: 816} & {Photo: 199} & {Cartoon: 324} & {Art painting: 201} \\
        {House} & \textbf{Photo: 280} &  {Art painting: 140} & {Sketch: 80} & {Cartoon: 288} \\
        {Person} & \textbf{Cartoon: 405} & {Sketch: 160} & {Photo: 202} & {Art painting: 449} \\
        \bottomrule 
    \end{tabular}
\end{table*}

\begin{table*}
    \centering
    \caption{Data split details of \textsl{unbalanced + flexible} setting on VLCS dataset when the domain ratio is 2:1:1. The dominant domain for each target domain is highlighted with the bold font.}
    \begin {tabular}{l|lll|l}
        \toprule
        {Class} & \multicolumn {3}{c|}{Source} & \multicolumn{1}{c}{Target} \\
        \midrule
        {0} & \textbf{Labelme: 56} & {Caltech: 27} &  {Sun: 14} & {Pascal: 231}  \\
        {1} & \textbf{Caltech: 86} & {Labelme: 43} & {Sun: 43}&  {Pascal: 489}  \\
        {2} & \textbf{Labelme: 62} & {Pascal: 30} & {Sun: 30} &  {Caltech: 83}  \\
        {3} & \textbf{Pascal: 294}  & {Caltech: 47}& {Labelme: 29} & {Sun: 21} \\
        {4} & \textbf{Pascal: 1049} & {Caltech: 524} & {Sun: 524} & {Labelme: 866} \\
        \bottomrule
    \end{tabular}
\end{table*}

\begin{table*}[ht]
    \centering
    \caption{Data split details of \textsl{unbalanced + flexible} setting on PACS dataset when the domain ratio is 1:1:1. The dominant domain for each target domain is highlighted with the bold font.}
    \begin{tabular}{l|lll|l}
        \toprule
        {Class} & \multicolumn{3}{c|}{Source} & \multicolumn{1}{c}{Target} \\
        \midrule
        {Dog} & Cartoon: 389 & {Art painting: 389} &  {Photo: 189} & {Sketch: 772}  \\
        {Elephant} & Cartoon: 457 & {Art painting: 255} & {Sketch: 457}&  {Photo: 202}  \\
        {Giraffe} & Photo: 182 & {Art painting: 182} & {Cartoon: 182} &  {Sketch: 753}  \\
        {Guitar} & Photo: 186  & {Art painting: 184}& {Sketch: 184} & {Cartoon: 135} \\
        {Horse} & Cartoon: 324 & {Photo: 199} & {Sketch: 324} & {Art painting: 201} \\
        {House} & Cartoon: 288 &  {Art painting: 288} & {Sketch: 80} & {Photo: 280} \\
        {Person} & Art painting: 449 & {Cartoon: 405} & {Photo: 432} & {Sketch: 160} \\
        \bottomrule 
    \end{tabular}
\end{table*}

\begin{table*}
    \centering
    \caption{Data split details of \textsl{unbalanced + flexible} setting on VLCS dataset when the domain ratio is 1:1:1. The dominant domain for each target domain is highlighted with the bold font.}
    \begin {tabular}{l|lll|l}
        \toprule
        {Class} & \multicolumn {3}{c|}{Source} & \multicolumn{1}{c}{Target} \\
        \midrule
        {0} & Labelme: 56 & {Caltech: 56} &  {Pascal: 56} & {Sun: 14}  \\
        {1} & {Labelme: 846} & {Caltech: 86} & {Sun: 652}&  {Pascal: 489}  \\
        {2} & {Pascal: 100} & {Caltech: 83} & {Labelme: 62} &  {Sun: 725}  \\
        {3} & {Pascal: 94}  & {Caltech: 47}& {Sun: 21} & {Labelme: 29} \\
        {4} & {Labelme: 866} & {Pascal: 866} & {Sun: 866} & {Caltech: 609} \\
        \bottomrule
    \end{tabular}
\end{table*}

\subsection{Details about the generation of MNIST-M in the setting of \textsl{composional + dominant + flexible + adversarial}}
The MNIST-M are generated by blending digit figures from the original MNIST dataset over patches extracted from images in BSDS500 dataset. The backgrounds are cropped from 200 images, resulting in 200 domains. The backgrounds from the same domain may be different given they are randomly cropped from the same image. We generate the adversarial setting by splitting the domains into 10 subsets responding to the classes. We randomly choose 1 subset for 1 class in the training data and choose 1 domain in the subset as the dominant domain. The ratio of the data from dominant domain to the data from other domains varies from 9.5:1 to 1:1. The subset chosen for one class for training is set to another class for testing, as well as the dominant domain.

\begin{figure*}[ht]
    \centering
    \includegraphics[width=0.8\textwidth]{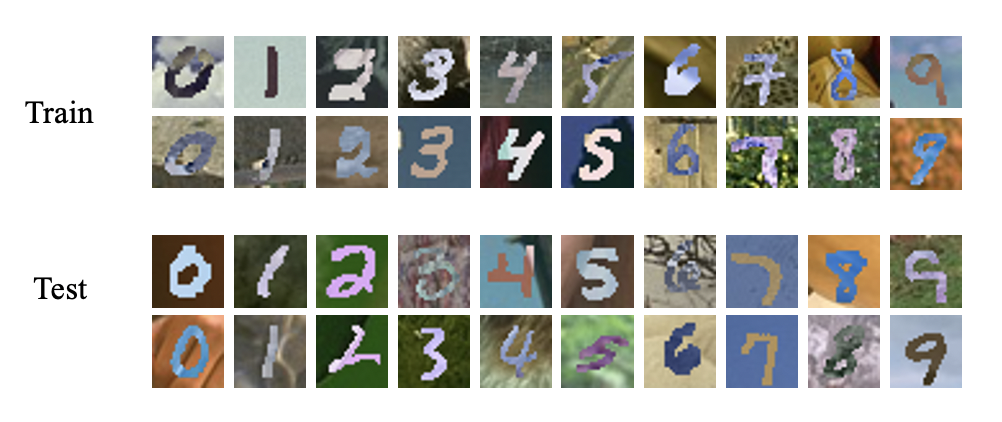}
    \caption{Example Images from MNIST-M}
    \label{fig:exmp_mnist}
\end{figure*}

\begin{figure*}[ht]
    \centering
    \includegraphics[width=0.8\textwidth]{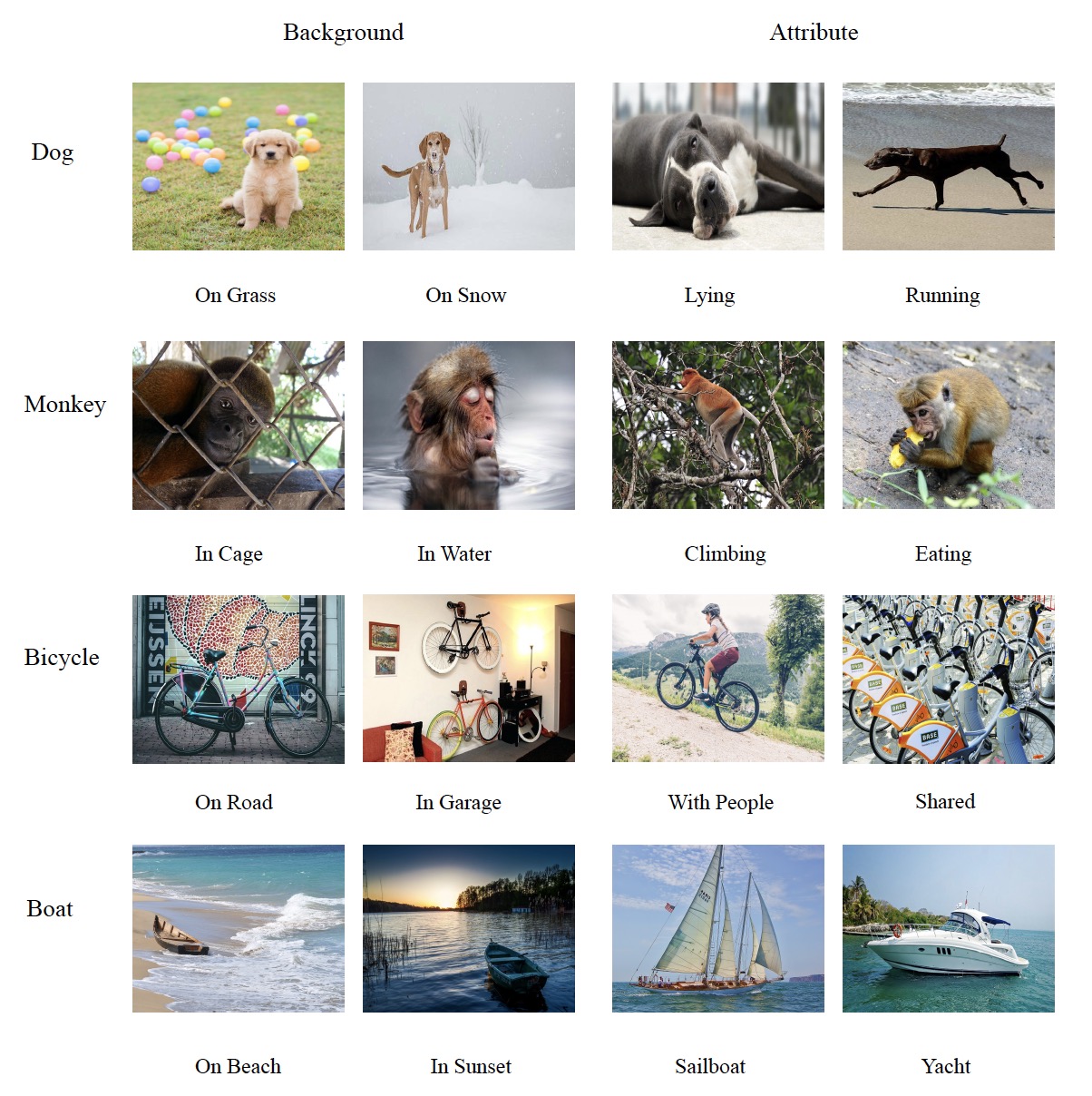}
    \caption{Example images with category and context labels from NICO}
    \label{fig:exmp_nico}
\end{figure*}

\subsection{NICO}
NICO is a dataset designed for distribution shifts problem. There are 19 categories and 10 contexts (domains) for each category. The domains for different category are various. The standard for split of contexts varies for different categories. For instance, some of the context are divided by the background of images such as `on water' or `on grass' while some by the posture of objects such as `running' or `standing'. Examples of images from NICO are shown in Figure \ref{fig:exmp_nico}.

There is a baseline method called CNBB in the original paper of NICO. We do not report the results of CNBB for the reason that it is designed for AlexNet and we fail to achieve reasonable results in our framework with CNBB. CNBB adopts Tanh function as the activation function and amplifies features from (-1, 1) to approach to \{-1, 1\} by a quantization loss shown as follows:

\begin{equation}
    \mathcal{L_{q}} = -\sum _{i=1}^{p} \left\lVert g_{\phi}(x_i) \right\rVert_{2}^{2}
\end{equation}

This loss harms ResNet significantly and it is hard to find proper hyperparameters for CNBB with ResNet as the backbone network. Hence, we do not report the results of CNBB.

\subsection{Examples of saliency maps}\label{app:saliency}
\begin{figure*}
    \centering
    \includegraphics[width=1.0\textwidth]{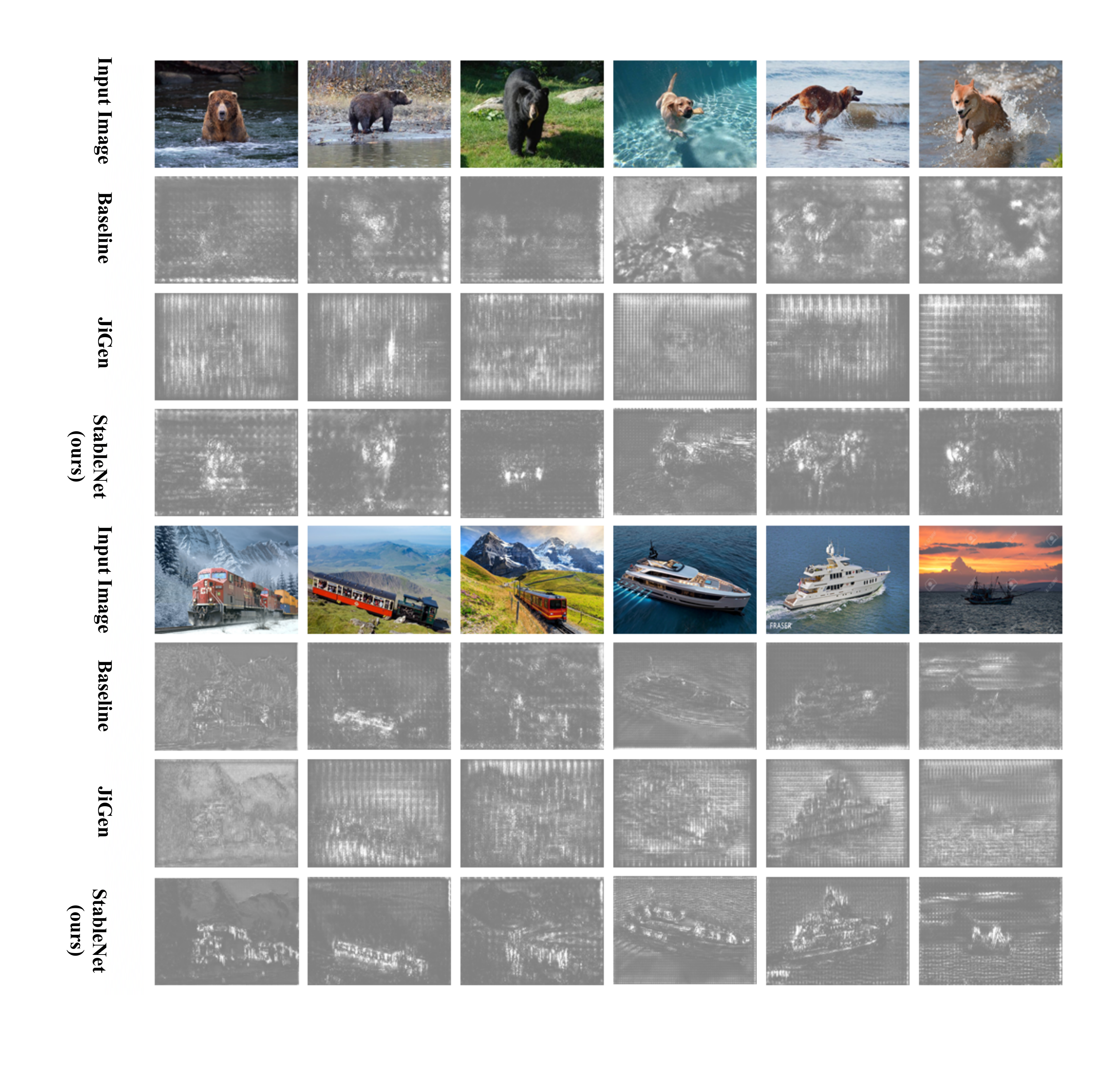}
    \caption{More saliency maps of the ResNet-18 model and StableNet.}
    \label{fig:saliency_more}
\end{figure*}

Examples of saliency maps are shown in Figure \ref{fig:saliency_more}.

The bright lines in saliency maps generated by JiGen demonstrates the effectiveness of the jigsaw puzzle, in which the model focuses more on the margins of any possible puzzles. And the highlight on the object in saliency maps generated by our method show that our model tends to focus on the object instead of the context. Therefore, our method help deep models learn the true connections between features and labels, resulting in models with stronger ability of generalization under distribution shifts.

\end{document}